%% file: main.tex
\documentclass[10pt,twocolumn,letterpaper]{article}

 \usepackage{iccv}              

\input{preamble}

\definecolor{iccvblue}{rgb}{0.21,0.49,0.74}
\usepackage[pagebackref,breaklinks,colorlinks,allcolors=iccvblue]{hyperref}

\def\paperID{8439} 
\def\confName{ICCV}
\def\confYear{2025}

\title{Realistic Clothed Human and Object Joint Reconstruction from a Single Image}

\author{Ayushi Dutta\textsuperscript{\thanks{Equal contribution}}
\and
Marco Pesavento\textsuperscript{\rm *}
\and
Marco Volino
\and
Adrian Hilton
\and
Armin Mustafa\\
Centre for Vision, Speech  and  Signal Processing (CVSSP), University of Surrey, Guildford, Uk\\
{\tt\small {\{a.dutta, m.pesavento, m.volino, a.hilton, a.mustafa\}}@surrey.ac.uk}
}

\begin{document}
\maketitle
\input{sec/0_abstract}
\input{sec/1_intro_A}
\input{sec/2_related_works}

\input{sec/3_methodology}

\input{sec/4_experiments}
\input{sec/5_conclusions}
\input{sec/X_suppl}
\clearpage
{
    \small
    \bibliographystyle{ieeenat_fullname}
    \bibliography{main}
}


\end{document}

%% file: preamble.tex
\usepackage[dvipsnames]{xcolor}
\definecolor{blue_mod}{RGB}{0,32,96}
\definecolor{red_mod}{RGB}{192,0,0}

\usepackage{breqn}
\usepackage{amsmath,bm}
\usepackage{algpseudocode}

\usepackage{enumitem}

\usepackage[dvipsnames]{xcolor}
\usepackage{gensymb}
\usepackage[]{multirow}
\usepackage{graphbox}
\usepackage{algorithm,stfloats}
\usepackage{soul}
\usepackage{mwe}
\usepackage{graphbox}
\usepackage{lipsum}
\usepackage{tabularx}
\usepackage{bm}
\usepackage[accsupp]{axessibility}
\usepackage{overpic}

\newcommand{\rone}[1]{\textcolor{blue}{R1}} 
\newcommand{\rtwo}[1]{\textcolor{magenta}{R2}} 
\newcommand{\rthree}[1]{\textcolor{teal}{R3}} 
 \usepackage{amssymb}
\newcommand{\name}{ReCHOR\xspace}
\newcommand{\dataname}{synHOR\xspace}

%% file: sec/0_abstract.tex
\begin{abstract}

Recent approaches to jointly reconstruct 3D humans and objects from a single RGB image represent 3D shapes with template-based or coarse models, which fail to capture details of loose clothing on human bodies. In this paper, we introduce a novel implicit approach for jointly reconstructing realistic 3D clothed humans and objects from a monocular view. For the first time, we model both the human and the object with an implicit representation, allowing to capture more realistic details such as clothing. This task is extremely challenging due to human-object occlusions and the lack of 3D information in 2D images, often leading to poor detail reconstruction and depth ambiguity. To address these problems, we propose a novel attention-based neural implicit model that leverages image pixel alignment from both the input human-object image for a global understanding of the human-object scene and from local separate views of the human and object images to improve realism with, for example, clothing details. Additionally, the network is conditioned on semantic features derived from an estimated human-object pose prior, which provides 3D spatial information about the shared space of humans and objects. To handle human occlusion caused by objects, we use a generative diffusion model that inpaints the occluded regions, recovering otherwise lost details. For training and evaluation, we introduce a synthetic dataset featuring rendered scenes of inter-occluded 3D human scans and diverse objects. Extensive evaluation on both synthetic and real-world datasets demonstrates the superior quality of the proposed human-object reconstructions over competitive methods.
\end{abstract}

%% file: sec/1_intro_A.tex
\vspace{-0.5cm}
\section{Introduction}
\label{sec:intro}
Realistic, personalized human avatars that seamlessly coexist with objects will shape the future of movies, games, telepresence, and the metaverse. The joint reconstruction of clothed humans and objects will be key to this vision. Emphasis will be on achieving realism in the reconstructed shapes to accurately reflect real-world characteristics. 
\begin{figure}[!t]
  \centering
\includegraphics[width=\linewidth]{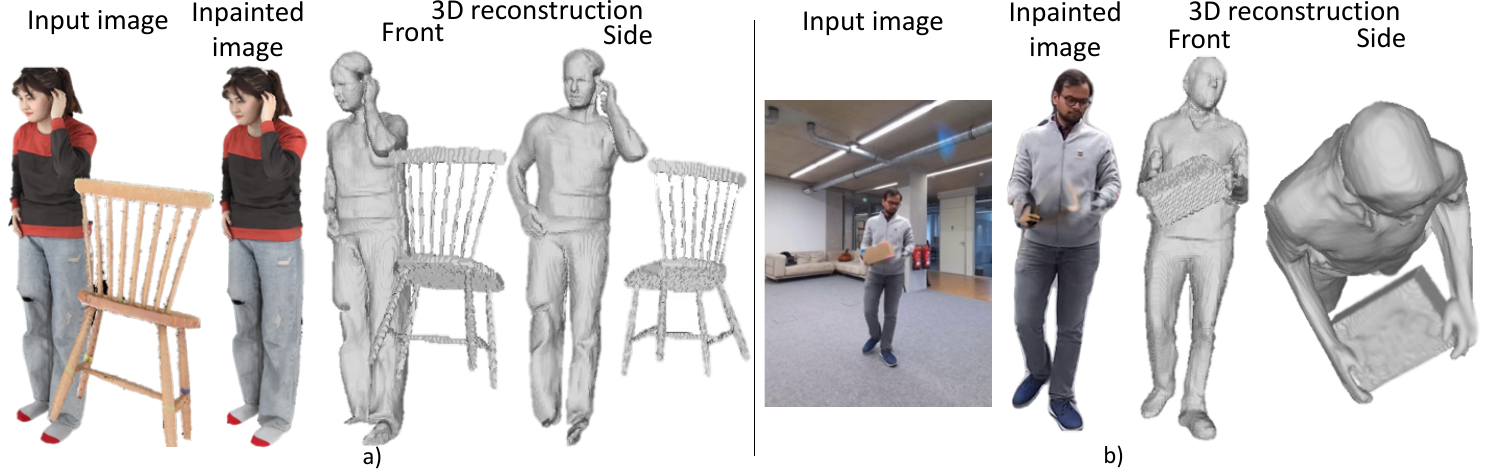}
\vspace{-7mm}
\caption{\name jointly reconstructs realistic clothed humans and objects from synthetic (a) and real (b) images by first handling human occlusion with a conditioned generative model followed by attention-based neural implicit model estimation.}
\label{fig:intro}
\vspace{-8mm}
\end{figure}
Motivated by this, we aim to jointly reconstruct realistic clothed humans and objects from a single-view human-object scene. This task presents significant challenges due to human-object occlusions and unknown camera parameters, which make it difficult to accurately infer the 3D spatial configuration (depth, scale, pose) as well as the realistic shape for reconstruction. Existing methods ~\cite{xie2022chore, zhang2020perceiving, nam2024contho, xie2023template_free, xie2023vistracker} that reconstruct 3D humans and objects from a single RGB image focus mainly on optimizing the 3D spatial configuration, failing to capture realistic details such as clothing, hairstyles, and the free-form geometry of human-object shapes. They represent 3D shapes using either parametric, template-based, or coarse models, which constrain the surface geometry, thereby limiting realism. We therefore explore implicit representations~\cite{mescheder2019occupancy, chen2019learning, park2019deepsdf}, which, unlike parametric models, enable realistic reconstruction without constraining the topology. However, while implicit representations can model realistic shapes and poses, they are prone to depth ambiguity if the only input is a 2D image without any explicit 3D depth information.
\\
To address these challenges, we propose \name, a novel framework for Realistic Clothed Human and Object joint Reconstruction. To obtain realistic surface details, \name incorporates a novel attention-based neural implicit model to estimate implicit representations of human-object shapes, assisted by a generative diffusion model that recovers details from regions of the human body occluded by the object. To resolve depth ambiguity, the neural implicit model is also conditioned on an estimated human-object pose prior, integrating 3D spatial information into the estimation process.
Specifically, we first segment the input human-object image to obtain separate human and object images. The generative diffusion model then inpaints the body regions occluded by the object in the human image, generating a full-body human image.  This image, along with the object image and additional inputs, form the 'local' context which embeds local details information. The input human-object image serves as the 'global' context, providing spatial cues between the human and the object. The neural implicit model uses pixel-aligned features from both image contexts, merging them through an attention-based architecture before estimating implicit representations, allowing the retrieval of realistic details while considering the contextual relationship between the human and the object.   
\\
Additionally, the neural implicit model is conditioned on semantic features derived from human-object pose priors, which are estimated using parametric human-object reconstruction methods~\cite{xie2022chore, zhang2020perceiving, nam2024contho, xie2023template_free}. These methods enforce geometric and spatial constraints to optimize for the 3D location, depth and scale of the human and object, providing \name with 3D spatial information essential to address the problem of depth ambiguity. The proposed model then computes the implicit representations in the reference frame of the predicted 3D location prior. 
By effectively decoupling depth ambiguity and detail surface retrieval, \name enables more accurate reconstruction of realistic clothed humans and objects, as shown in~\cref{fig:overview}.
\\
Due to the lack of datasets with high-quality 3D ground-truth human-object scenes, we create \dataname, a synthetic dataset for Human Object Reconstruction to train and evaluate \name. We 
generate several 3D spatial configurations of human-object scenes by randomly placing 3D human scans from THuman2.0~\cite{yu2021function4d} with selected object meshes from BEHAVE~\cite{bhatnagar2022behave} and HODome~\cite{zhang2023neuraldome}.  We also evaluate our model on the real-world BEHAVE~\cite{bhatnagar2022behave} dataset and demonstrate superior performance of reconstructions against state-of-the-art methods.
Our key contributions include: 
\begin{itemize}
  \item A novel framework to jointly reconstruct realistic clothed humans and objects from single images. This is the first work that represents realistic human details in the joint reconstruction of a non-parametric human-object shape.
  \item A novel attention-based neural implicit network to estimate the implicit representation of realistic clothed humans and objects. Pixel-aligned features are extracted from local and global views and then merged along with 3D spatial information via transformer encoders, capturing realistic details while learning contextual information across local and global scenes.
  \item We demonstrate superior reconstruction quality compared to the state-of-the-art methods, both quantitatively and qualitatively, on synthetic and real datasets. 
\end{itemize}

%% file: sec/2_related_works.tex
\section{Related works}
\label{sec:related}
\noindent{\textbf{3D human and object reconstruction:}
To reconstruct 3D human and object jointly, previous methods use parametric models to fit human and object meshes satisfying various constraints. PHOSA~\cite{zhang2020perceiving} and D3D-HOI~\cite{xu2021d3d} each proposed an optimization based framework with physical constraints on scale and predefined contact priors. 
Wang \emph{et al.}~\cite{wang2022reconstructing} modeled 3D human-object shapes from an image using commonsense knowledge from large language models. Holistic++~\cite{chen2019holistic++} modeled fine-grained human-object relations in a scene using Markov chain Monte Carlo method.  CHORE proposed to fit a parametric model to a learned neural-implicit functions. Vistracker~\cite{xie2023vistracker}, InterTrack~\cite{xie2024intertrack} reconstruct human-object from a single video, by specifically modeling the temporal context.
Recently, CONTHO~\cite{nam2024contho} proposed a method to refine human-object reconstruction from an image by 3D guided contact estimation.
ProciGen~\cite{xie2023template_free} proposed a Hierarchical Diffusion Model to reconstruct human and object. None of the current approaches for single-view human-object reconstruction can recover realistic clothed humans at the same fidelity as \name. Alternate methods~\cite{sun2021neural, jiang2022neuralhofusion, jiang2023instant, zhang2023neuraldome} that can reconstruct clothed human-object, either use sparse, multi-views or monocular RGBD image as inputs, thereby relaxing their constraints and are thus not comparable to our work.}
%

\noindent{\textbf{Neural implicit model for single-view 3D reconstruction:}
Early reconstruction methods cannot produce realistic 3D shapes due to the discretized nature of traditional representations like voxels, meshes or point clouds. The introduction of the implicit representation for 3D reconstruction~\cite{mescheder2019occupancy, chen2019learning,park2019deepsdf} has led deep learning approaches to adopt this continuous representation because it can represent fine details on the reconstructed shapes. These works estimate the implicit representation with neural implicit models.
Several works have since used neural implicit models for object reconstruction from single images. Early models estimate occupancy~\cite{chen2019learning, mescheder2019occupancy} or signed distance fields~\cite{park2019deepsdf,xu2019disn,zhang2021generalized} using multi-layer perceptrons conditioned on input features. Recent works improve the object reconstruction by incorporating prior knowledge into implicit functions ~\cite{liu2019learning, chen2023g2ifu};
using monocular geometric cues ~\cite{yu2022monosdf}; combining explicit templates with implicit representations~\cite{fahim2022enhancing, wang2023ifkd} or leveraging input global and local features ~\cite{li2021d2im,arshad2023list}.
Other implicit models focus on reconstructing high-quality 3D human shapes from single images.  PIFu~\cite{pifu} introduced the pixel-aligned implicit function to retrieve detailed human shapes. Building on this, several works improved quality using normal maps~\cite{pifuhd, icon} or super-resolution shapes~\cite{surs}; addressed depth ambiguity using parametric models~\cite{pamir, geopifu}; achieved complete reconstructions with diffusion models~\cite{google1, google2, ho2024sith} or incorporating additional data as depth maps~\cite{econ, anim} or unconstrained images~\cite{cosmu}.
These works cannot jointly reconstruct human-object shapes since they are category-specific, with human-focused methods unable to reconstruct objects and vice-versa. \name jointly reconstructs realistic clothed humans and objects from a single image. }

%% file: sec/3_methodology.tex
\section{Realistic 3D shapes of clothed humans and objects}
\label{sec:method}
\begin{figure*}
  \centering
\includegraphics[width=\linewidth]{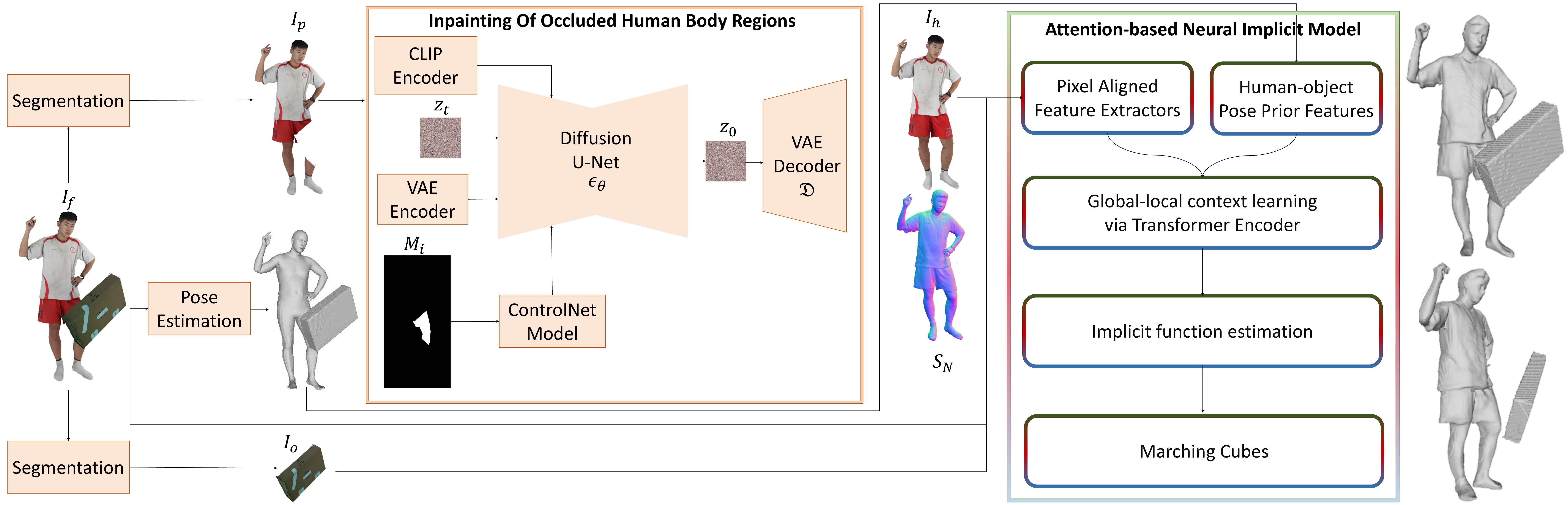}
\vspace{-8mm}
\caption{\name overview: Given an input image of a human-object scene, we first use a generative model to inpaint occluded human body regions, guided by a mask of missing areas and the segmented input of the human. Next, the generated image, along with an estimated normal map, the input image, the segmented object image, and estimated pose parameters, are processed by an attention-based neural implicit model. This model jointly estimates the implicit representation of the human-object shape.}
\label{fig:overview}
\vspace{-5mm}
\end{figure*}
We introduce \name, a novel framework for Realistic Clothed Humans and Objects Reconstructions from a single RGB image depicting both the human and the object. In the proposed pipeline, shown in~\cref{fig:overview}, the image is first segmented to separate the human from the object, and pose parameters of the SMPL-H~\cite{smplh} model and object are estimated. Regions of the human body that  are occluded by the object in the input image are inpainted using the generative power of an image-conditioned diffusion model. Implicit representations of both the human $s_h$ and the object $s_o$ are then estimated using a novel attention-based neural implicit model that incorporates the input RGB image $I_f$, the generated full-body human image $I_h$, the object image $I_o$ and the estimated pose parameters. Compared to related works, our approach jointly reconstructs realistic clothed humans and objects while avoiding depth ambiguity between them.
\subsection{Inpainting of occluded human body regions}
\label{ssec:diff_model}
This work addresses cases in images where an object occludes the human. When such occlusion happens, regions of the human body are missing in the input image, leaving the neural implicit model without the information needed to estimate these regions. We propose to leverage the generative capability of diffusion models to inpaint the occluded body regions, using the partial human image $I_p$ along with a mask of the occluded regions $M_i$. Given the input image $I_f$, we first apply semantic segmentation to separate the human from the object, while estimating the SMPL-H~\cite{smplh} and object pose. Using the SMPL-H model, we then generate a mask of the human $M_s$, and we obtain the human-object intersection mask $M_i$ as $M_i=M_s-M_p$, where $M_p$ is the segmented partial human mask.
\\Inspired by SiTH~\cite{ho2024sith}, we leverage an image-condition latent diffusion model~\cite{rombach2022high} (LDM) to learn the conditional distribution of the missing human body regions. Rather than training the LDM from scratch, we adopt a fine-tuning strategy~\cite{kumari2023multi, zhang2023adding} that optimizes the cross-attention layers of a pretrained diffusion U-Net~\cite{rombach2022high}. The learning is conditioned using a ControlNet~\cite{zhang2023adding}, which processes $M_i$ to guide the generation. Additionally, we condition the U-Net with features from the partial human image $I_p$, extracted via a pre-trained CLIP~\cite{clip} image encoder and a VAE encoder $\varepsilon$,  to ensure that the generated image $I_h$ matches the appearance of $I_p$. These features, along with randomly sampled noise $\epsilon$, are input to the LDM model $\epsilon_{\theta}$, conditioned by the ControlNet, which generates a latent code $z$ within the VAE latent distribution $z = \varepsilon(I^{gt}_h)$, where $I^{gt}_h$ is the ground-truth full-body human image. The output full-body image $I_h$ is then generated by decoding with the VAE decoder  $\mathcal{D}$  a latent code $\tilde{z}$ derived through iterative denoising of Gaussian noise: $I_h = \mathcal{D}(\tilde{z})$.
\\We define the objective function for fine-tuning as:
\vspace{-0.2cm}
\begin{equation}
    \min_{\theta}\mathbb{E}_{z \sim \varepsilon(I^{gt}_h),t, \epsilon \sim \mathcal{N}(\textbf{0},\textbf{I}) }\left\lVert \epsilon - \epsilon_{\theta}(z_t, t, I_p, M_i) \right\rVert^2_2
\vspace{-0.2cm}
\end{equation}
The ground-truth latent code $z_0 = \varepsilon(I^{gt}_h)$ is diffused over 
$t$ time steps, resulting in the noisy latent $z_t$. The image-conditioned LDM model $\epsilon_{\theta}$ predicts the noise $\epsilon$ added to the noisy latent $z_t$, based on time step $t \sim [0, 1000]$ and conditional inputs $I_p$ and $M_i$.
\\To inpaint the occluded body regions during inference, we generate a latent $\tilde{z}_0$ by starting with Gaussian noise $z_T \sim \mathcal{N}(\textbf{0},\textbf{I})$ and using an iterative denoising process. The final full-body image $I_h$ is  obtained with the decoder $\mathcal{D}$:
\vspace{-0.2cm}
\begin{equation}
    I_h=\mathcal{D}(\tilde{z}_0)=\mathcal{D}(f_{\theta}(z_T,I_h,M_i))
\vspace{-0.2cm}
\end{equation}
where $f_{\theta}$ indicates the iterative denoising process of $\epsilon_{\theta}$.
Finally, a 2D normal map $S_N$ is estimated from $I_h$ and concatenated with it. For simplicity, we continue to refer to this concatenated result as $I_h$ throughout the paper.
\begin{figure}
  \centering
\includegraphics[width=\linewidth]{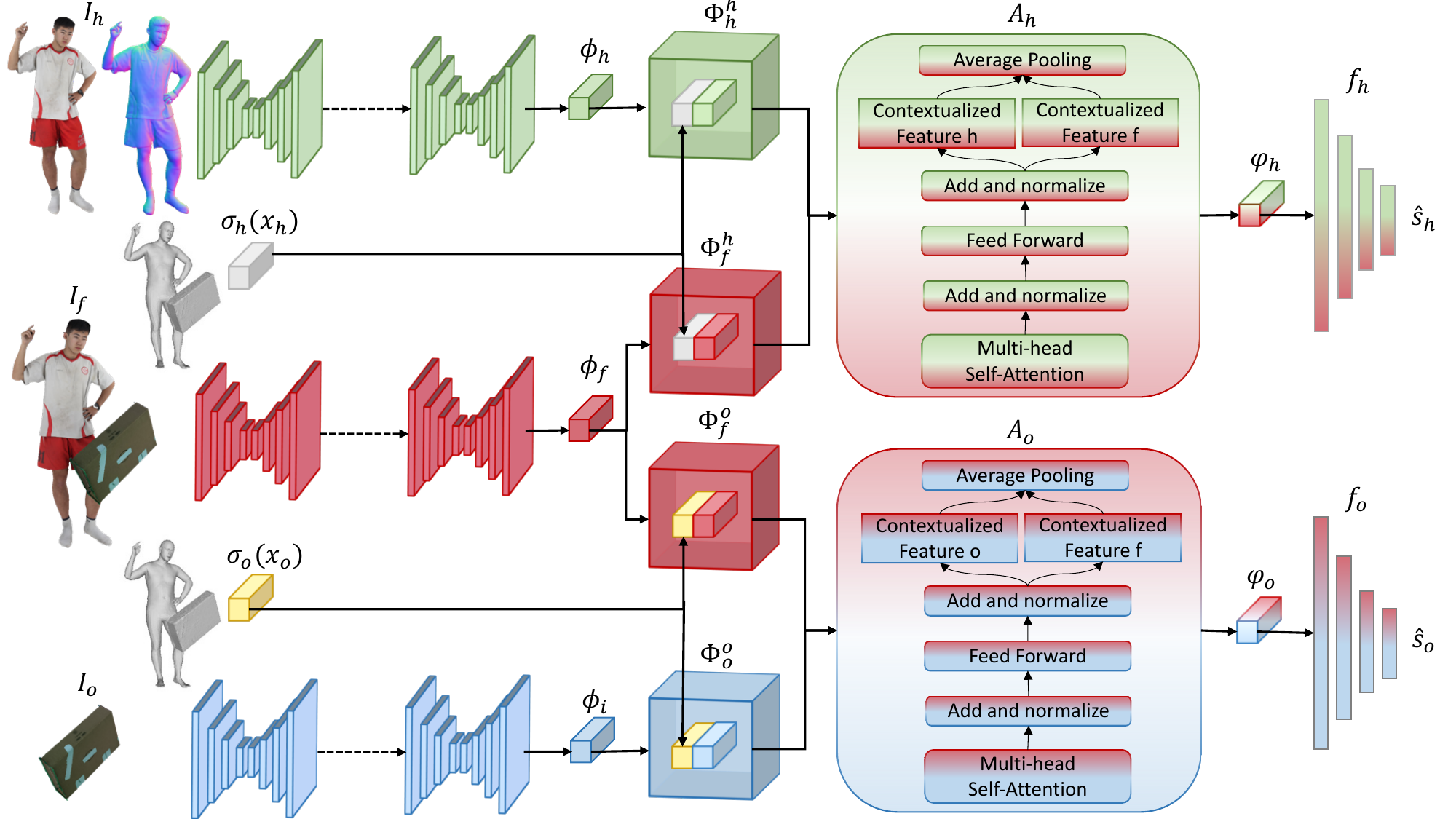}
\vspace{-8mm}
\caption{The attention-based neural implicit model first extracts pixel-aligned features from the input images to capture local details. It then uses a transformer encoder to merge these features, learning global and local contextual information about the scene. Finally, the model estimates the implicit representations for both humans and objects. Human-object pose priors provide 3D spatial information to address depth ambiguity.}
\label{fig:fig_module}
\vspace{-6mm}
\end{figure}
\subsection{Attention-based neural implicit model}
\label{ssec:neural}
To reconstruct the 3D shapes of both humans and objects, we introduce a novel attention-based neural implicit model that jointly estimates their implicit representations. This model aims to estimate an implicit representation that defines a surface as the level set of a function $f$, \ie $f(X) = 0$ where $X$ is a set of 3D points in $\mathbb{R}^3$.  The reconstructed surface is then defined as the zero level-set of $f$:
\vspace{-0.2cm}
\begin{equation}
    f' = \{ x:\; f(x) = 0,\; x\in\mathbb{R}^3\}
\vspace{-0.2cm}
\end{equation}
The proposed implicit model comprises  of multiple modules, as shown in 
\cref{fig:fig_module}.
\\\textbf{Pixel-aligned feature extractors}: Previous methods on 3D human reconstruction~\cite{pifu,pifuhd,surs} demonstrate that projecting a 3D point $x \in \mathbb{R}^3$ in the embedded image feature space $\phi(I)$ extracted with a convolutional stacked hourglass network significantly increases the quality of 3D human shapes. \name first extracts a feature embedding for each input image  $\phi_{\{h, o, f\}}(I_{\{h, o, f\}})$. A first set of points $X_h$ is then projected onto $I_f$ and $I_h$ via perspective projection $\pi$ and linked to the corresponding features $\phi_h$ and $\phi_f$ to obtain pixel-alignment with the human: $\Phi_{\{h,f\}}^h=\phi_{\{h,f\}}(\pi(X_h,I_{\{h,f\}}))$. Similarly, the pixel-aligned object features $\Phi_{\{o,f\}}^o=\phi_{\{o,f\}}(\pi(X_o,I_{\{o,f\}}))$ are obtained by projecting a different set of points $X_o$ on $I_o$ and $I_f$ and indexing them with the  corresponding features $\phi_o$ and $\phi_f$.
\\\textbf{Human-object pose prior features:} Multiple 3D spatial configurations of humans and objects can project to the same 2D image, leading to difficulties in estimating their position in 3D space and depth-scale ambiguity. Since parametric model-based human-object reconstruction methods optimize 3D location, depth, and scale using geometric and spatial constraints, we leverage them to address depth-scale ambiguity and anchor the 3D spatial location. We condition our model on semantic features of the SMPL-H and object template estimated by the parametric model-based method, and we compute the neural representations of both human and object relative to the predicted SMPL-H center.  The object position is defined relative to this center, to learn its 3D spatial relationship with the human. To define the human pose features, for a query point $x_h \in X_h$, we look for the closest point $x_h^*$ on SMPL-H, \ie $x_h^* = \arg \min_{x_h^p} || x_h - x_h^p||$ where $x_h^p$ are points on the SMPL-H mesh. Human pose prior features $\sigma_h(x_h) = [d_h, v_h, z_h]$ comprise three elements: a signed distance value $d_h$ between $x_h$ and $x_h^*$, a visibility label $v_h \in \{1, 0\}$ where $v_h$ indicates if $x_h^*$ is visible in the image when SMPL-H and the object mesh is projected together, and a relative depth-aware feature $z_h = (x_h^z - z_c)$ where $z_c$ is the depth of the SMPL-H center and $x_h^z$ is the depth of the query point $x_h$. Analogously, for a query point $x_o \in X_o$ and its closest point on the object mesh, $x_o^*$, the object pose prior features $\sigma_o(x_o) = [d_o, v_o, z_o]$ are three elements: a signed distance value $d_o$ between $x_o$ and $x_o^*$, a visibility label $v_o \in \{1, 0\}$ and a relative depth-aware feature $z_o = (x_o^z - z_c)$ where $x_o^z$ is the depth of the query 
point $x_o$. 
\\\textbf{Global-local context learning via transformer encoders}: As shown in~\cref{ssec:abl}, simply concatenating the features of the human $\Phi_h^h$ and object $\Phi_o^o$ with those of the input image $\Phi_f^{\{h,o\}}$ deteriorates network performance, making it less robust to noise of the input image. Additionally, the network's ability to integrate and reason about features across the entire image is limited by the receptive field of convolutional layers.  To better understand the relationship between the human and the object in the space, we propose using two attention-based encoders, $A_o$ and $A_h$ that merge the feature extracted by the feature extractor module. An attention score is computed for the two paired images by assessing the compatibility between a query and its corresponding key, producing two feature embeddings:  one for the human $\varphi_h=A_h(\Phi_h^h,\Phi_f^h,\sigma_h(x_h))$ and another for the object $\varphi_o=A_o(\Phi_o^o,\Phi_f^o,\sigma_o(x_o))$. The pose features $\sigma_{\{h,o\}}$ obtained in the previous module are also concatenated to integrate spatial information. Each feature $\varphi_{\{h,o\}}$ integrates the information from its corresponding local image $I_{\{h,o\}}$, combined with the global context of the input image $I_f$. A more comprehensive understanding of the scene is obtained by contextualizing the global and local information. As a result, the human feature $\varphi_{h}$ contains detailed information from the human image, as well as contextual information of the global scene from the input image. Similarly, the object feature $\varphi_{o}$ integrates scene-level information from $I_f$.
\\\textbf{Implicit function estimation}: The implicit functions of the human $f_h$ and object $f_o$ are modeled with two separate multi-layer perceptrons (MLPs), which jointly estimate the occupancy values of  $X_h$ and $X_o$:
\vspace{-0.2cm}
\begin{equation}
    \begin{aligned}
\hat{s} = \begin{cases} \hat{s}_h=f_h(\varphi_h, X_h), \quad \hat{s}_h \in \mathbb{R},  \\ \hat{s}_o=f_o(\varphi_o, X_o), \quad \hat{s}_o \in \mathbb{R} \end{cases}
\end{aligned}
\vspace{-0.2cm}
\end{equation}
where $\hat{s}$ is the implicit representation of the final human-object shape, with $\hat{s}_h$ representing the human component and $\hat{s}_o$ representing the object component.
\\The neural implicit model is trained end-to-end with the following objective function $\mathcal{L}=\mathcal{L}_{h}+\mathcal{L}_{o}$:
\vspace{-0.2cm}
\begin{equation}
    \mathcal{L}_{h}=\frac{1}{N_h}\sum_{j=1}^{N_h}\left|f_{h}(\varphi^j_h, X^j_h)-f_{h}^{gt}(\varphi^j_h X^j_h)\right|^2
\vspace{-0.2cm}
\end{equation}
where $N_h$ is the number of points $X_h$ and $f_{h}^{gt}$ is the ground-truth implicit function for the human shape, and 
\vspace{-0.2cm}
\begin{equation}
    \mathcal{L}_{o}=\frac{1}{N_o}\sum_{j=1}^{N_o}\left|f_{o}(\varphi^j_o, X^j_o)-f_{o}^{gt}(\varphi^j_o, X^j_o)\right|^2
\vspace{-0.2cm}
\end{equation}
where $N_o$ is the number of points $X_o$ and $f_{o}^{gt}$ is the ground-truth implicit function for the object shape. 
\\\textbf{Inference}: During inference, an input RGB image $I_f$, containing both a human and an object, is first processed with semantic segmentation to generate separate masks for human $M_p$ and object $M_o$. Human and object poses are also estimated from $I_f$. If the object partially occludes the human, the segmented human image $I_p$ and the intersection mask $M_i$, are processed by the diffusion module. This results in a full-body image $I_h$ from which the 2D normal map $S_N$ is estimated. These outputs, along with the input image $I_f$, the object image $I_o$ and the pose features $\sigma_{\{h,o\}}$ are then processed by the attention-based neural implicit model. This model estimates the occupancy $\hat{s}$ of a set of random 3D points $X$ of the 3D space. Finally, the 3D shape is obtained by extracting iso-surface $f = 0.5$ of the probability field $\hat{s}$ at threshold 0.5 with the Marching Cubes algorithm \cite{lorensen1987marching}.

%% file: sec/4_experiments.tex
\section{Experiments}
\label{sec:experiments}
\textbf{Datasets:} Due to the lack of 3D human-object datasets with high-quality meshes that represent clothing details, we introduce the dataset \dataname to train our model. \dataname is a synthetic dataset created using $526$ 3D human scans from Thuman 2.0~\cite{yu2021function4d} and $27$ 3D object scans including all 20 objects from BEHAVE~\cite{bhatnagar2022behave} and selected 7 from HODome~\cite{zhang2023neuraldome}. We picked random pairs of human and object meshes, where the object was initialized with a random pose and optimized to be in contact with the human. We then simulated 6 random translations of the human-object pair within the FOV of a perspective camera placed at the origin, and rendered 180 views for each translation. Finally, we discard the images where the object is not in view.  We train \name with 500 subjects of \dataname. For quantitative evaluation, we use 99 images from \dataname. We test the generalization power of \name by inferring the human-object shapes from real images taken from BEHAVE~\cite{bhatnagar2022behave}. BEHAVE is a real-world dataset that captures interactions between 8 humans and 20 objects. Due to its real-world setting, high-quality ground truth meshes are unavailable; therefore, we perform only a qualitative evaluation.
\\\textbf{Implementation details.} We use SAM~\cite{kirillov2023segment} to segment the input image into object and human and create the masks $M_p$ and $M_i$. CHORE~\cite{xie2022chore} is applied for pose estimation of the human and object. Once the diffusion model generates the inpainted image, we segment the occluded body region with $M_i$ and merge it into the partial human image $I_p$ to obtain the full-body human image $I_h$. The 2D normal map is estimated from $I_h$  using pix2pixHD~\cite{wang2018high} as in PIFuHD~\cite{pifuhd}.
To train the attention-based neural implicit model, we sample two sets of $N=200000$ points around the ground-truth object and human surfaces, using a mix of uniform and importance sampling with variances $\sigma=0.06, 0.01, 0.035$. From these, we randomly select $N_h=N_o=20000$ points to create the $X_h$ and $X_o$ subsets. These points are projected into the input images following the Kinect camera model from the BEHAVE dataset~\cite{bhatnagar2022behave}. Both the diffusion and neural implicit models are trained on a single A100 GPU.
See supplementary for additional details about implementation.
\\\textbf{Metrics:} We follow the evaluation
in SiTH~\cite{ho2024sith} to compute 3D metrics point-to-surface distance (P2S),
Chamfer distance (CD), normal consistency (Normal), Intersection over Union (IoU) and fScore~\cite{tatarchenko2019single} on the generated meshes. The metrics are computed between the combined 3D human-object ground truth and the estimated reconstruction, with the estimated shape aligned to the ground truth by rescaling and translating it to match the human centroid.
\subsection{Comparisons}
\label{ssec:comp}
\vspace{-1mm}
Our goal is to demonstrate that \name can jointly reconstruct realistic clothed human and object shapes from single images. To the best of our knowledge, \name is the first framework that represents realistic human details in the joint reconstruction using an unconstrained topology for the human-object shape since related methods rely on template-based or coarse representations, reducing reconstruction realism. We also compare \name with approaches that focus solely on high-quality human reconstruction to highlight their inability to reconstruct objects.
\begin{figure*}[h]
  \centering
\includegraphics[width=\linewidth]{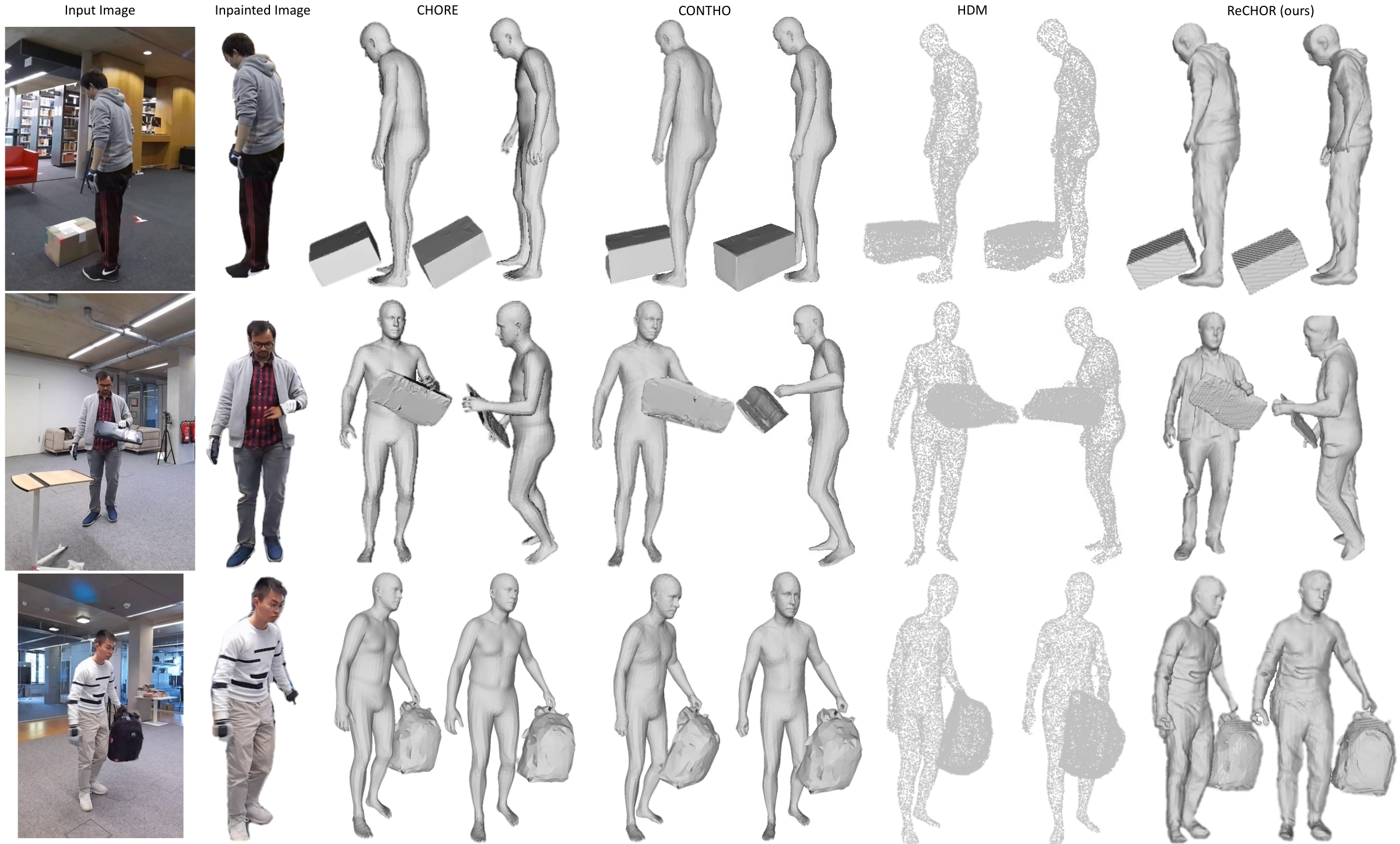}
\vspace{-7mm}
\caption{Qualitative evaluations against methods which aim to reconstruct human-object jointly with examples from BEHAVE dataset~\cite{bhatnagar2022behave}. Note that HDM generates point clouds rather than meshes. Front and side views are shown.}
\label{fig:comp2}
\vspace{-6mm}
\end{figure*}

\noindent{\textbf{Parametric human-object reconstruction methods}: We compare \name against state-of-the-art human-object reconstruction methods from a single image that use a template-based or coarse representation of the 3D shapes, namely CHORE~\cite{xie2022chore}, CONTHO~\cite{nam2024contho} and HDM~\cite{xie2023template_free}. As shown in ~\cref{fig:comp2}, \name is the only method that reconstructs realistic clothed human shapes along with the objects. Related works use a parametric-based representation of humans, which lacks detail and significantly reduces realism. These approaches are designed solely to predict the human-object spatial configuration. \name leverages them as priors and significantly improves the realism of the reconstruction while preserving the 3D spatial configuration. Note that we do not quantitatively evaluate parametric methods as the quality of their human shapes is much lower than non-parametric shapes.} 
\begin{table}[t]
\centering
\resizebox{\linewidth}{!}
{\input{tables/comp}}
\vspace{-3.5mm}
\caption{Quantitative comparisons on \dataname dataset between \name and related works that reconstruct high-quality 3D shape in the bottom part. Results from comparisons with baselines created from PIFuHD~\cite{pifuhd} are also presented.}
\vspace{-6mm}
\label{tab:comp}
\end{table}

\begin{figure*}[t]
  \centering
\includegraphics[width=\linewidth]{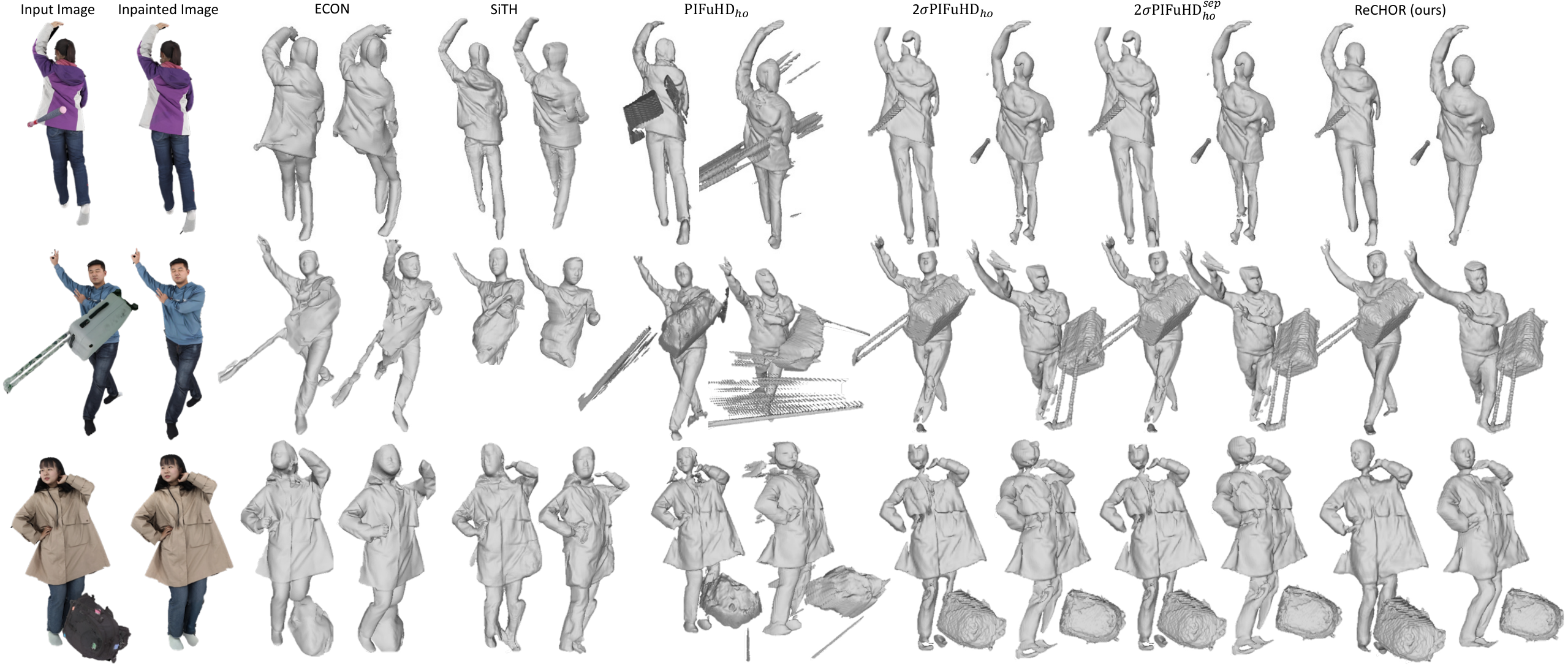}
\vspace{-7mm}
\caption{Visual comparisons from \dataname dataset with approaches that aim to reconstruct 3D humans as well as with baselines designed for fair comparisons. Front and side views are shown.}
\label{fig:comp1}
\vspace{-6.5mm}
\end{figure*}

\noindent{\textbf{Realistic human reconstruction methods}: We evaluate \name against methods designed for high-quality human shape reconstruction from single images using neural implicit models (PIFuHD~\cite{pifuhd}, ECON~\cite{econ}, SiTH~\cite{ho2024sith}). To demonstrate that these works cannot jointly reconstruct human and object shapes, we use the RGB image $I_f$ as input. We also retrain PIFuHD on synHOR ($\mathrm{PIFuHD}_{\mathrm{ho}}$) and introduce several baseline models based on the PIFuHD architecture to showcase \name's superiority in joint human-object reconstruction. We first use two MLPs instead of one to estimate an implicit representation of both human and object ($2\mathrm{PIFuHD}_{\mathrm{ho}}$). Since PIFuHD does not rely on 3D spatial information, we repeat the previous experiments by concatenating the human-object pose feature $\sigma$ to the extracted features (indicated with the prefix $\sigma$). Features are then extracted from only segmented human and object images ($2\sigma\mathrm{PIFuHD}^{sep}_{\mathrm{ho}}$) and finally, the images ($I_f, I_h, I_o$) are concatenated together as input before feature extraction ($2\sigma\mathrm{PIFuHD}^{all}_{\mathrm{ho}}$). We do not  repeat these experiments with ECON and SiTH because adapting their methods to our dataset would require substantial modifications. As shown in~\cref{tab:comp}, \name achieves the best quantitative results and, as seen in~\cref{fig:comp1}, is the only approach that can jointly reconstruct realistic clothed human and object shapes without noise. Human-focused approaches (ECON, SiTH, PIFuHD) struggle with object reconstruction, impacting human shape quality as well. Retraining PIFuHD with \dataname dataset ($\mathrm{PIFuHD}_{\mathrm{ho}}$) and even adding two MLPs and pose parameters ($2\sigma\mathrm{PIFuHD}_{\mathrm{ho}}$) is insufficient to achieve the realism of \name, which outperforms all methods significantly. See supplementary for more comparisons.}
\vspace{-1mm}
\subsection{Ablation Study}
\label{ssec:abl}
\begin{figure*}[t]
  \centering
\includegraphics[width=\linewidth]{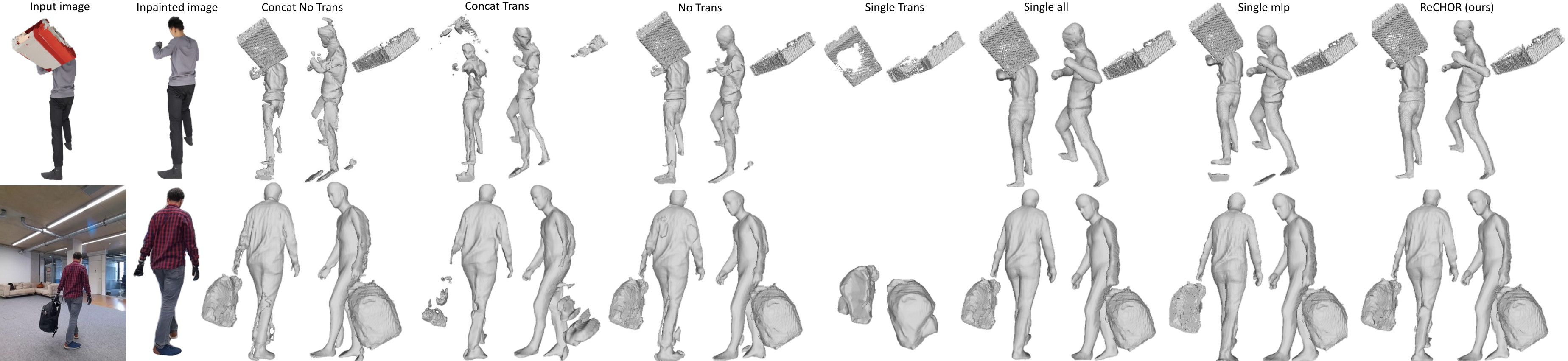}
\vspace{-7mm}
\caption{Qualitative results showing the effect of using different configurations of the attention-based neural implicit model. Front and side views are shown from \dataname in the top row, from BEHAVE~\cite{bhatnagar2022behave} in the bottom.}
\label{fig:abl1}
\vspace{-4mm}
\end{figure*}

\vspace{-1mm}
\indent \textbf{Effectiveness of architecture configuration}: We demonstrate the significant improvement achieved by using the architecture configuration  explained in~\cref{ssec:neural}. 6 alternate configurations of the architecture of the  attention-based neural implicit model of \name are trained and tested:
\begin{itemize}[noitemsep]
\item\textbf{Single MLP}: To prove the effectiveness of estimating an implicit representation for each human and object, a single MLP is used on concatenated features $\varphi_{\{h,o\}}$ to estimate a single occupancy $\hat{s}$ for both human and object. 
\item\textbf{Single Trans}: Use of only a single transformer encoder $A$ to merge human, object and input image features. $\hat{s}_h$ and $\hat{s}_o$ are estimated from the merged feature using $f_h$ and $f_o$.
\item\textbf{Single All}: Combination of the previous two configurations with a single transformer encoder and a single MLP.
\item\textbf{No Trans}: The embeddings $\phi_{\{h,o,f\}}$ from the feature extractors are simply concatenated without applying the transformer encoders, highlighting the benefit of using self-attention. $\hat{s}_h$ and $\hat{s}_o$ are estimated.
\item\textbf{Concat Trans}: Instead of extracting features directly from $I_f$, this configuration concatenates  $I_f$ with both $I_h$ and $I_o$. $\phi_f$ is not extracted. $\phi_o$ and $\phi_h$ are merged with a transformer encoder before estimating $\hat{s}_h$ and $\hat{s}_o$.
\item\textbf{Concat No Trans}: Similar to \textbf{Concat Trans}, but  $\phi_o$ and $\phi_h$ are processed separately by  $f_h$ and $f_o$ respectively.
\end{itemize}
Both quantitative (\cref{tab:abl1}) and qualitative (\cref{fig:abl1}) results prove that the \name's architecture outperforms the other configurations. Specifically, replacing two MLPs with one (\textbf{Single MLP}) reduces reconstruction accuracy  and detail, as seen in the hand of the BEHAVE model. 
If two MLPs process the same feature obtained by merging all the features with a single transformer (\textbf{Single Trans}), human-specific information is lost, resulting in an incomplete human reconstruction.  Using both a single MLP and a single transformer degrades quality, producing smoother surfaces than \name. Omitting the transformer encoder altogether (\textbf{No Trans}) makes the network less robust and introduces noise in the reconstruction, showing the importance of global-local contextualization. Finally, concatenating the input image $I_f$ directly with $I_o$ and $I_h$, significantly reduces quality compared to \name, both when a transformer is used to merge human-object features (\textbf{Concat Trans}) and when it is not (\textbf{Concat No Trans}), proving that extracting features directly from $I_f$ improves performance. 
\begin{figure*}[h]
  \centering
\includegraphics[width=0.85\linewidth]{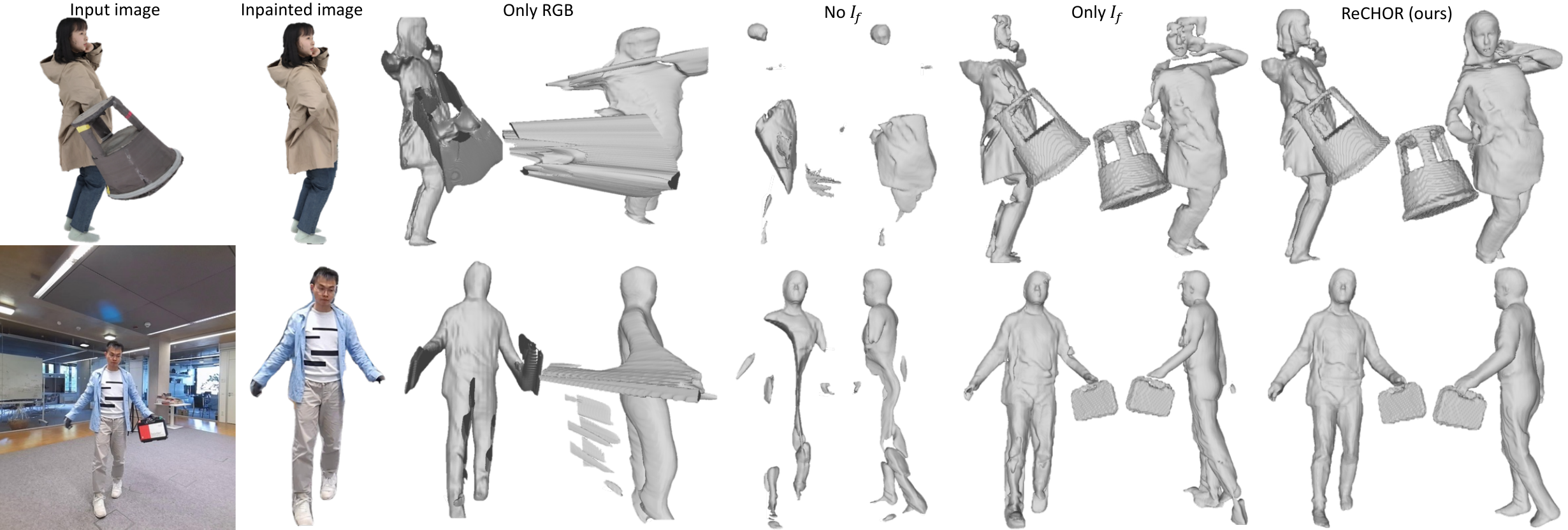}
\vspace{-3mm}
\caption{Qualitative results showing the effect of using different combinations of input data. Front and side views are shown from \dataname in the top row, and on BEHAVE~\cite{bhatnagar2022behave} in the bottom row.}
\label{fig:abl2}
\vspace{-6mm}
\end{figure*}
\begin{table}[t]
\vspace{-3mm}
\centering
\resizebox{\linewidth}{!}
{\input{tables/abl1}}
\vspace{-3.5mm}
\caption{Quantitative results on \dataname dataset obtained by modifying the architecture of the network.}
\vspace{-6.5mm}
\label{tab:abl1}
\end{table}
\indent \textbf{Effect of different inputs:}
\name uses multiple inputs to reconstruct the 3D human-object, including the RGB image $I_f$, full-body human image $I_h$, 2D surface normal $S_N$, segmented object $I_o$ and human-object poses $\sigma_{\{h,o\}}$. In this ablation study, we show the advantages of combining all these inputs. We adapt \name architecture to process different combinations of input data. In each configuration, two MLPs are still used to estimate $\hat{s}_h$ and $\hat{s}_o$.
\begin{itemize}[topsep=0pt,partopsep=0pt,itemsep=0pt,parsep=0pt] 
    \item\textbf{No} $\mathbf{S_N}$: Same  as \name but without  2D normal map $S_N$ concatenated to $I_h$.    
    \item\textbf{No} $\boldsymbol{\sigma_{\{h,o\}}}$:  Same as \name but without  processing the human-object pose features $\sigma_{\{h,o\}}$.
    \item\textbf{Only RGB}: Same as \name without 2D normal map $S_N$ and human-object pose features $\sigma_{\{h,o\}}$.
    \item\textbf{No} $\mathbf{I_f}$: No input image $I_f$, so the image features $\Phi_f^{\{h,o\}}$  are not extracted. The human image feature $\Phi_h^{h}$ is merged with the object one $\Phi_o^{o}$ with a transformer encoder.
    \item\textbf{No} $\mathbf{I_o}$: No object image $I_o$. The human image feature $\Phi_h^{h}$ is merged with the input image feature $\Phi_f^{h}$ using $A_h$.
    \item\textbf{No} $\mathbf{I_h}$: No full-body human image $I_h$. The object feature $\Phi_o^{o}$ is merged with the input image feature $\Phi_f^{o}$ using $A_o$.
    \item\textbf{Only} $\mathbf{I_f}$: Only the input image $I_f$ is processed. No transformer is applied.
\end{itemize}
\begin{table}[t]
\centering
\resizebox{\linewidth}{!}
{\input{tables/abl2}}
\vspace{-3.5mm}
\caption{Quantitative results on \dataname dataset  obtained with different combinations of input data.}
\vspace{-6mm}
\label{tab:abl2}
\end{table}
The benefit of incorporating all input data types is demonstrated by the superior quantitative results in~\cref{tab:abl2}. This is further supported by the visual results in~\cref{fig:abl2}, where more realistic and accurate human-object shapes are obtained with \name. When the input RGB image is omitted (\textbf{No } $\mathbf{I_f}$), the model lacks global scene understanding, resulting in severe artifacts. Likewise, if human and object images are not used (\textbf{Only}  $\mathbf{I_f}$), the network lacks local information about them. \textbf{Only RGB} shows the importance of using both normal maps and pose features to reproduce realistic details and avoid depth ambiguity in the reconstruction. Additional ablations are in the supplementary. 
\\\textbf{Limitations and Future work:} Modeling realistic clothed human and object interactions is an extremely challenging problem. 
This work represents the first step toward achieving this goal. The accuracy of the interaction between humans and objects will be further improved in future works. We also aim to improve the quality of the object shape by introducing higher quality ground-truth object shapes in the dataset to learn high-frequency details for the objects as well.  While our approach specifically addresses cases where a human is occluded by an object, we will extend it to handle object occlusions as well. 
We also aim to retrieve textures for clothed human-object reconstructions. 

%% file: tables/comp.tex
\begin{tabular}{c|ccccc}
                                                                  & \multicolumn{5}{c}{\textbf{synHOR}}                                                          \\ \hline
Methods                                                           & P2S $\downarrow$ & CD $\downarrow$ & IoU $\uparrow$ & Normal $\uparrow$ & f-Score $\uparrow$ \\ \hline
PIFuHD                                                            & 33.75            & 67.59           & .4555          & .6595             & 18.07              \\
ECON                                                              & 36.48            & 79.39           & .3952          & .6547             & 14.57              \\
SiTH                                                              & 31.88            & 68.88           & .3371          & .6521             & 14.33              \\ \hline
$\mathrm{PIFuHD}_{\mathrm{ho}}$                                   & 43.50            & 91.97           & .3917          & .5661             & 11.00              \\
$\sigma\mathrm{PIFuHD}_{\mathrm{ho}}$                             & 31.66            & 50.78           & .4228          & .7090             & 20.59              \\
$2\mathrm{PIFuHD}_{\mathrm{ho}}$                                  & 48.29            & 80.09           & .3482          & .5347             & 10.79              \\
$2\sigma\mathrm{PIFuHD}_{\mathrm{ho}}$                            & 28.82            & 36.83           & .6966          & .7196             & 25.62              \\
\multicolumn{1}{c|}{$2\sigma\mathrm{PIFuHD}^{sep}_{\mathrm{ho}}$} & 25.83            & 33.20           &     .8176           & .7204             & 29.17              \\
\multicolumn{1}{l|}{$2\sigma\mathrm{PIFuHD}^{all}_{\mathrm{ho}}$} & 23.87          & 26.53           & .8172          & .7601             & 32.72              \\
\name (ours)                                       & \textbf{14.23}   & \textbf{17.15}  & \textbf{.8902} & \textbf{.8241}    & \textbf{48.84}    
\end{tabular}

%% file: tables/abl1.tex
\begin{tabular}{c|ccccc}
                            & \multicolumn{5}{c}{\textbf{synHOR}}                                                          \\ \hline
Methods                     & P2S $\downarrow$ & CD $\downarrow$ & IoU $\uparrow$ & Normal $\uparrow$ & f-Score $\uparrow$ \\ \hline
Concat No Trans             & 25.59            & 30.38           & .8246          & .7324             & 29.66              \\
Concat Trans                & 48.75            & 84.18           & .4057          & .5440             & 9.459              \\
No Trans                    & 22.90            & 24.66           & .8447          & .7683             & 33.97              \\
Single All                  & 15.05            & 18.12           & .8752          & .8248             & 47.71              \\
Single Trans                & 52.78            & 101.1           & .2284          & .5406             & 7.103              \\
Single MLP                  & 17.57            & 24.70           & .8267          & .7986             & 43.98              \\
\name (ours) & \textbf{14.23}   & \textbf{17.15}  & \textbf{.8902} & \textbf{.8241}    & \textbf{48.84}    
\end{tabular}

%% file: tables/abl2.tex
\begin{tabular}{c|ccccc}
                            & \multicolumn{5}{c}{\textbf{synHOR}}                                                          \\ \hline
Methods                     & P2S $\downarrow$ & CD $\downarrow$ & IoU $\uparrow$ & Normal $\uparrow$ & f-Score $\uparrow$ \\ \hline
Only RGB                    & 47.86            & 78.25           & .4472          & .5962             & 11.33              \\
No $\sigma_{h,o}$           & 44.14            & 71.72           & .5064          & .6175             & 12.85              \\
No $S_N$                    & 14.24            & \textbf{17.12}  & .8661          & .8024             & 41.46              \\
Only $I_f$                  & 28.82            & 36.83           & .6966          & .7196             & 25.62              \\
No $I_f$                    & 45.80            & 100.0           & .3473          & .5284             & 5.76               \\
No $I_h$                    & 53.24            & 108.4           & .2239          & .5374             & 7.66               \\
No $I_o$                    & 51.77            & 91.32           & .3516          & .5535             & 8.88               \\
\name (ours) & \textbf{14.23}   & 17.15           & \textbf{.8902} & \textbf{.8241}    & \textbf{48.84}    
\end{tabular}

%% file: sec/5_conclusions.tex
\vspace{-1mm}
\section{Conclusion}
\label{sec:conclusion}
\vspace{-1mm}
We presented \name, the first novel method that can jointly reconstruct realistic clothed humans and objects from a human-object scene image. \name combines the power of attention-based neural implicit learning with a generative diffusion model and human-object pose conditioning to produce realistic clothed human and object 3D shapes. Extensive ablations on various network architecture components of \name demonstrate the effectiveness of the proposed approach. Our experiments show that our method generalizes well to a real world dataset on which it was not trained and outperforms the state-of-the-art methods in reconstructing realistic human-object scenes.

%% file: sec/X_suppl.tex
\clearpage
\setcounter{page}{1}
\maketitlesupplementary
\section{Overview}
In the following sections of the supplementary material we present:
\begin{itemize}
    \item Additional details about the implementation of \name (\cref{sec:impl});
    \item A table containing a list of symbol used in the main paper is presented in~\cref{sec:notations}
    \item Additional ablation studies to demonstrate the effectiveness of the proposed framework (\cref{sec:abl_suppl});
    \item Additional results of \name and related works on both \dataname and BEHAVE~\cite{bhatnagar2022behave} datasets (\cref{sec:comp_supp});
    \item A more detailed discussion about results, limitations and future works (\cref{sec:discussion});
\end{itemize}
\section{Implementation details}
\label{sec:impl}
\subsection{Diffusion model}
Inspired by SitH~\cite{ho2024sith}, our image-conditioned latent diffusion model~\cite{rombach2022high} used to generate missing body regions as described in \cref{ssec:diff_model}, is trained by fine-tuning the diffusion U-Net’s~\cite{rombach2022high} weights. These weights are initialized using the Zero-1-to-3~\cite{lugaresi2019mediapipe} model, combined with a trainable ControlNet~\cite{zhang2023adding} model, following the default network setups with input channel adjustment. The ControlNet input is a 1 channel mask and the diffusion U-Net input is a 3-channel RGB image of size 512 x 512. We train the models with a batch size of 6 images on a single A100 NVIDIA GPU, with a learning rate of $4\times10^-6$ and adopting a constant warmup scheduling. The ControlNet model’s conditioning scale is fixed at 1.0. We use classifier-free guidance in our training, involving a dropout rate of 0.05 for the image-conditioning. During inference, a classifier-free guidance scale of 2.5 is applied to generate the output images.
\begin{table}[t!]
\centering
\resizebox{\linewidth}{!}
{\input{tables/notation}}
\vspace{-3mm}
\caption{List of notations used in the main paper.}
\label{tab:notation}
\end{table}
\subsection{Attention-based neural implicit model}
We detail the implementation of our attention-based neural implicit model described in \cref{ssec:neural}. For each input image of size 2048x1536, we take a crop around the human-object bounding box center and resize it to 512px for the network. Following PIFu~\cite{pifu}, we use a four-stack Hourglass network that yields a 256-dimensional feature map for querying pixel-aligned features. A standard multi-head transformer-style architecture is applied for the two attention-based encoders  $A_h$ and $A_o$, as shown in~\cref{fig:fig_module}. Given three vectors query $Q=M_q\phi$, key $K=M_k\phi$ and value $V=M_v\phi$ as the embedding of the original feature $\Phi$ and parameterized by matrices $M_q$, $M_k$ and $M_v$, an attention score is computed for each input feature $\Phi^h_{\{h,f\}}$ for $A_h$ and $\Phi^o_{\{o,f\}}$ for $A_o$ based on the compatibility of a query with a corresponding key:
\begin{equation}
    Attention(Q,K,V)=softmax\left(\frac{QK^T}{\sqrt{d_k}}\right)V
\end{equation}
where $d_k$ is the common dimension of $K$, $Q$ and $V$. Multiple heads are used to compute features for the human and object:
\begin{equation}
\begin{gathered}
    MultiHead(Q, K, V)=concat(H_1, ..., H_h)W^o
    \\
    H_i=Attention(QW^q_i, KW^k_i,V W^v_i)
\end{gathered}
\end{equation}
where $QW^q_i$, $KW^k_i$,$V W^v_i$  are the parameters of $Q$, $K$ and $V$, and $W^o$o the parameters of the final projection. The final feature is computed as the mean of the processed features with each merged feature $\varphi_{\{h,o\}}$ containing local information from the human or object image along with global scene information from the input image. Finally, to estimate the neural implicit representations, each decoder $f_h$, $f_o$ is an MLP with the number of neurons (259, 1024, 512, 256, 128, 1) and skip connections at 2nd, 3rd and 4th layers. We train the network end-to-end using Adam optimizer with a learning rate of $1e-4$ and batch size 4. 
\section{Notations}
\label{sec:notations}
To facilitate the understanding of the main paper,~\cref{tab:notation} presents a list of the notations used in the paper.
\section{Additional ablation studies}
\label{sec:abl_suppl}
\subsection{Effect of different inputs}
\begin{figure*}[b]
  \centering
\includegraphics[width=\linewidth]{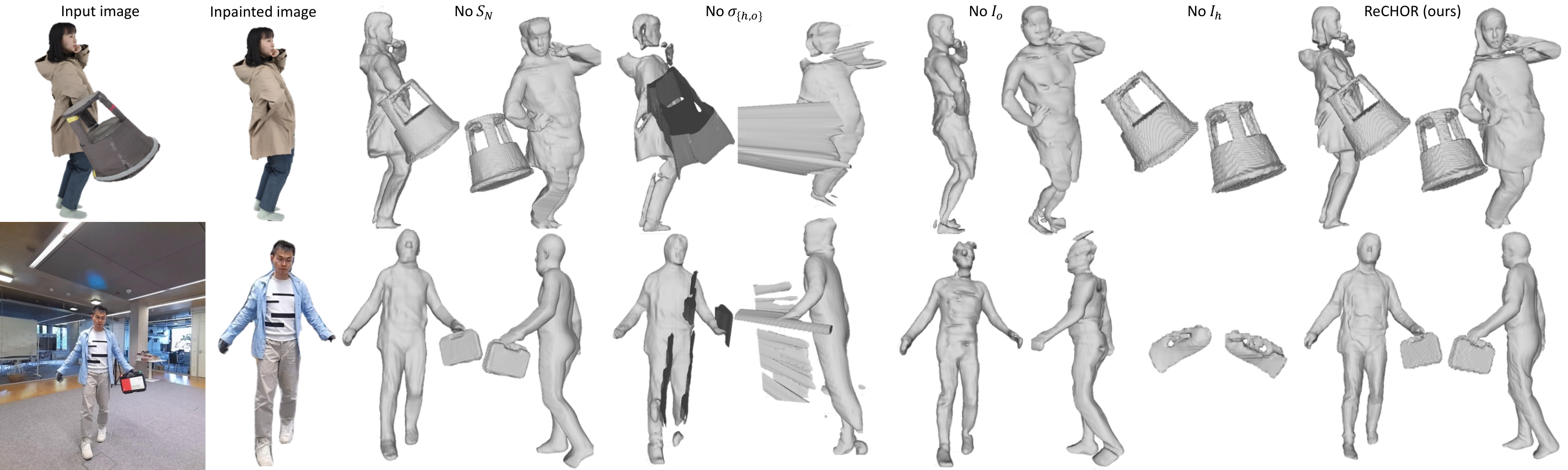}
\vspace{-6mm}
\caption{Qualitative results showing the effect of using different configurations of the attention-based neural implicit model with methods that have not been shown in~\cref{fig:abl2} of the main paper. Front and side views are shown from \dataname in the top row, from BEHAVE~\cite{bhatnagar2022behave} in the bottom.}
\label{fig:abl2_supp}
\end{figure*}

\cref{fig:abl2} of the main paper shows examples of human-object shapes reconstructed using different combinations of input. Additional results from the configurations analyzed in the second ablation study of~\cref{ssec:abl} and not included in~\cref{fig:abl2} are shown in~\cref{fig:abl2_supp}. 
The highest quality human-object shapes are obtained using \name, confirming what was proved in~\cref{ssec:abl}. Smoother shapes are reconstructed when normal maps are excluded (No $S_N$) while omitting pose features (No $\sigma_{{h,o}}$) introduces depth ambiguity. Excluding the full-body human input (No $I_h$) or the object input (No $I_o$) prevents the network from reconstructing the human or object, respectively.
\subsection{Training configurations}
\begin{table}[h]
\centering
\resizebox{\linewidth}{!}
{\input{tables/train_conf}}
\vspace{-3mm}
\caption{Quantitative results obtained with different training configurations of \name.}
\label{tab:train_config}
\end{table}
\begin{figure*}[t]
  \centering
\includegraphics[width=\linewidth]{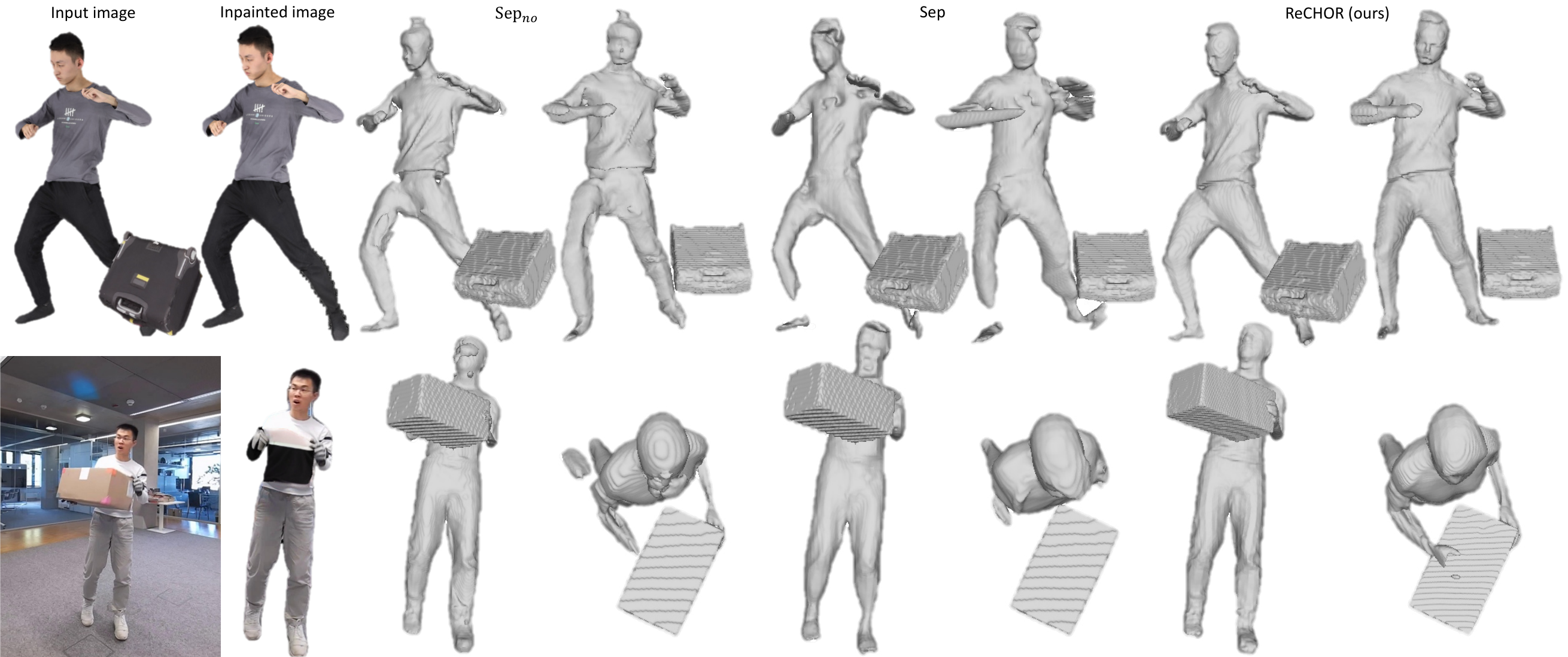}
\caption{Examples of results obtained with different training configurations of \name. Front and side views are shown from \dataname in the top row, from BEHAVE~\cite{bhatnagar2022behave} in the bottom.}
\label{fig:train_config}
\end{figure*}
\noindent The proposed attention-based neural implicit model is trained end-to-end to enable better joint reasoning about the human-object scene. This section presents results obtained by modifying \name's training configuration:
\begin{itemize}
    \item $\mathbf{Sep_{no}}$: Two separate networks are trained. One reconstructs the human shape using the full-body human image generated by the inpainting diffusion model and the relative human pose feature. The other reconstructs the object shape using the object image and pose feature. No transformer encoder is used.
    \item $\mathbf{Sep}$: Similar to $\mathbf{Sep_{no}}$, two networks are trained separately to estimate the human and object implicit representations. However, in this configuration, the input RGB image is also considered. Each network extracts features from the input RGB image and merges them with either the full-body human or object features using transformer encoders.
\end{itemize}
As demonstrated by the quantitative evaluation in~\cref{tab:train_config} and the qualitative results (\cref{fig:train_config}), training the attention-based neural implicit model end-to-end significantly improves performance. Both $\mathbf{Sep}$ and $\mathbf{Sep_{no}}$ training configurations result in less accurate human-object reconstructions. Training separate networks for humans and objects leads to reconstruction errors in interaction regions. This demonstrates how the proposed design can embed contextual information about global and local scenes, learning spatial relationship between human and object.
\begin{figure*}[t]
  \centering
\includegraphics[width=0.85\linewidth]{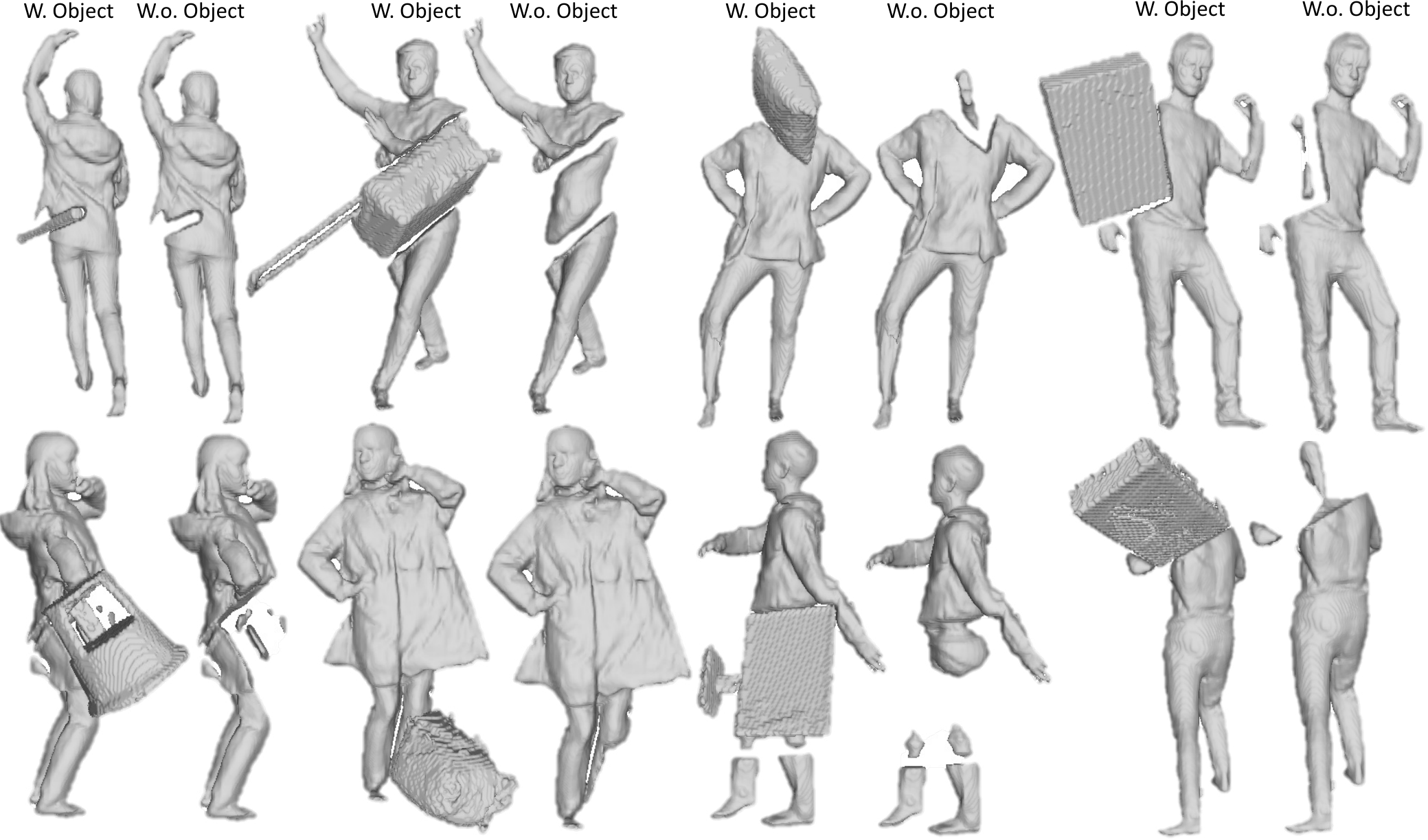}
\vspace{-4mm}
\caption{Effect of not applying the inpainting module of \name before estimating the implicit representation. The same examples used in the paper are illustrated with (w.) and without (w.o) the object.}
\label{fig:no_inp}
\end{figure*}
\subsection{Inpainting module}
\begin{figure*}[h]
  \centering
\includegraphics[width=\linewidth]{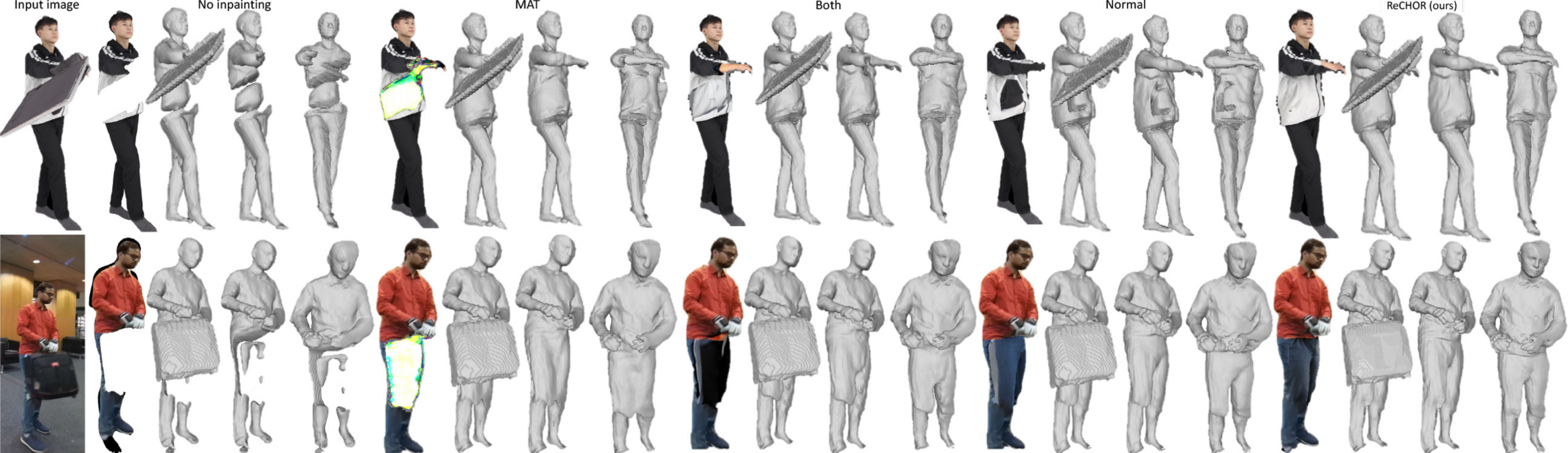}
\vspace{-6mm}
\caption{Human-object shapes reconstructed by changing the diffusion module of \name. Front view with and without the object and side views without the object are shown from \dataname in the top row, from BEHAVE~\cite{bhatnagar2022behave} in the bottom.}
\label{fig:mesh_inp}
\end{figure*}

\begin{table}[h]
\centering
\resizebox{\linewidth}{!}
{\input{tables/inp}}
\vspace{-3mm}
\caption{Quantitative evaluation of changing the diffusion module of \name. The human-object shapes reconstructed with the attention-based neural implicit module of \name are used for evaluation.}
\label{tab:inp}
\end{table}
\begin{figure*}[t]
  \centering
\includegraphics[width=\linewidth]{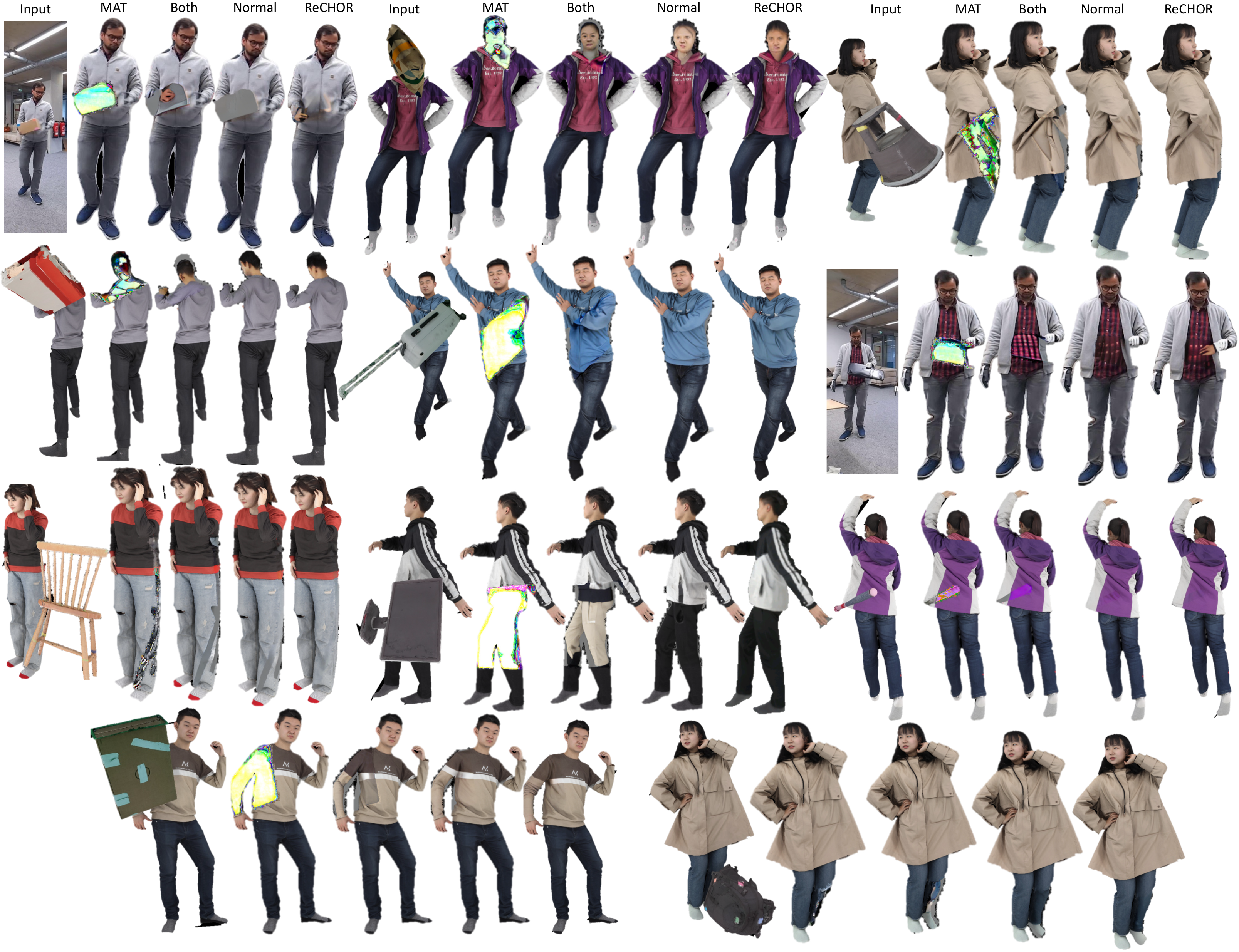}
\vspace{-6mm}
\caption{Visual comparisons of inpainted images generated using different inpainting approaches.}
\label{fig:rgb_inp}
\end{figure*}
\noindent In scenes containing both humans and objects, the object often occlude parts of the human body. When only partial regions of the body are visible, the network fails to reconstruct the occluded regions. 
This limitation is shown in~\cref{fig:no_inp}, where \name is applied without the inpainting module to generate a full-body human image. Instead, the partial human image is directly input into the attention-based neural implicit model, which fails to reconstruct these occluded regions, producing holes in the human shapes. Inpainting the missing body regions in the input image is crucial to reconstruct realistic clothed human and object shapes.
\\\name addresses the challenge of human occlusion by leveraging the generative capability of diffusion models to inpaint the occluded body regions. A fine-tuning strategy is adopted to optimize the cross-attention layers of a pre-trained diffusion model, conditioning it on the mask of the body regions requiring inpainting. In this section, we explore different solutions for inpainting, including MAT~\cite{li2022mat}, a transformer-based model for large hole inpainting, conditioning the diffusion model with the SMPL~\cite{smplh} normal map (Normal), and conditioning with both the mask and the normal map (Both). 
The effect of using these approaches for the inpainting module of \name is quantitatively evaluated by estimating the implicit representation of human-object shapes using their generated full-body human image (\cref{tab:inp}). Examples of human-object reconstructions using these methods for inpainting are shown in~\cref{fig:mesh_inp} while~\cref{fig:rgb_inp} illustrates the input images shown in the paper and generated using the discussed inpainting methods.
\\As expected, the worst results occur when no inpainting is applied (No inpainting). MAT struggles to produce realistic body regions in the RGB images, propagating noise into the final reconstruction. Conditioning the diffusion model with both the SMPL normal map and the mask also degrades performance. The best results are achieved when either the SMPL normal or the mask of the occluded regions is processed using ControlNet. In particular, processing only the mask proves more robust, as seen in the top-left example of~\cref{fig:rgb_inp} and in the reconstructed human-object shape in the top row of~\cref{fig:mesh_inp}, where unnatural black regions in the full-body human image generated with Normal configuration caused gaps in the reconstructed shape.
\\Note that we do not quantitatively evaluate the generated images against ground truth, since the main goal of \name is 3D reconstruction rather than inpainting.
\section{Additional visual comparisons}
\label{sec:comp_supp}
\begin{figure*}[b!]
  \centering
\includegraphics[width=0.99\linewidth]{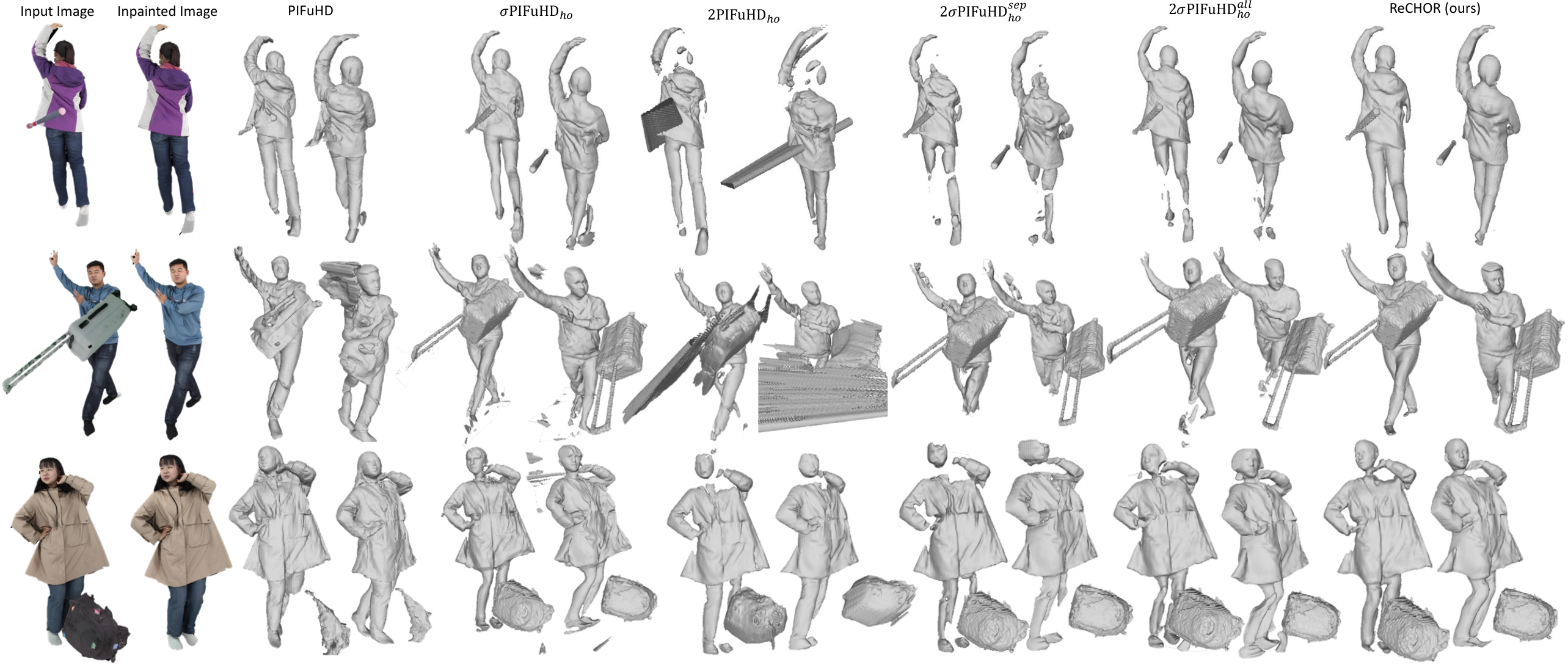}
\vspace{-4mm}
\caption{Visual comparisons from \dataname with related works not shown in~\cref{fig:comp1} of the main paper. Front and side views are shown.}
\label{fig:comp1_supp}
\end{figure*}

\begin{figure*}[t]
  \centering
\includegraphics[width=\linewidth]{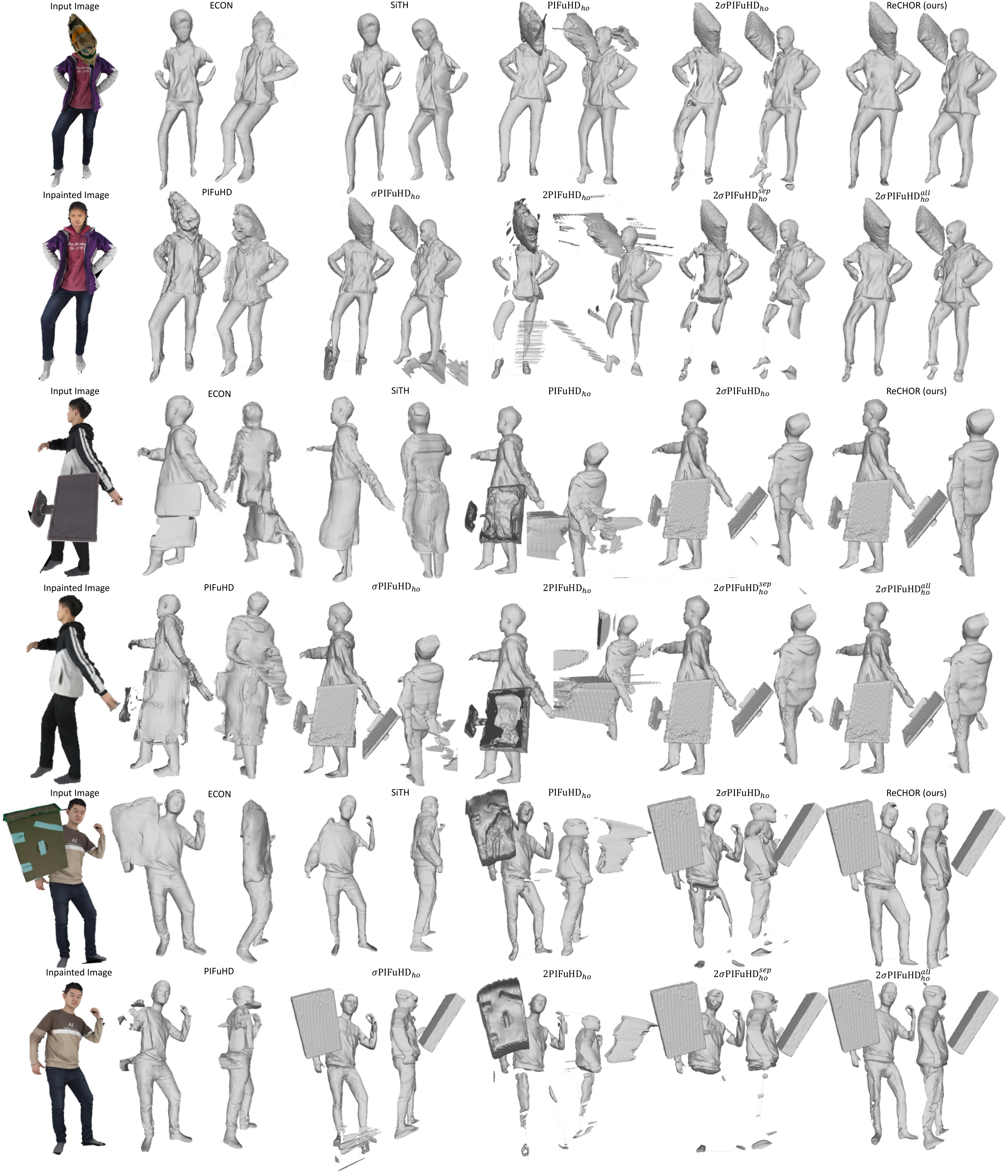}
\vspace{-6mm}
\caption{Additional visual results from \dataname. Examples obtained with methods that aim to reconstruct 3D humans as well as with the baselines considered in the main paper are shown. Front and side views are shown.}
\label{fig:thuman_supp}
\end{figure*}

\begin{figure*}[t]
  \centering
\includegraphics[width=\linewidth]{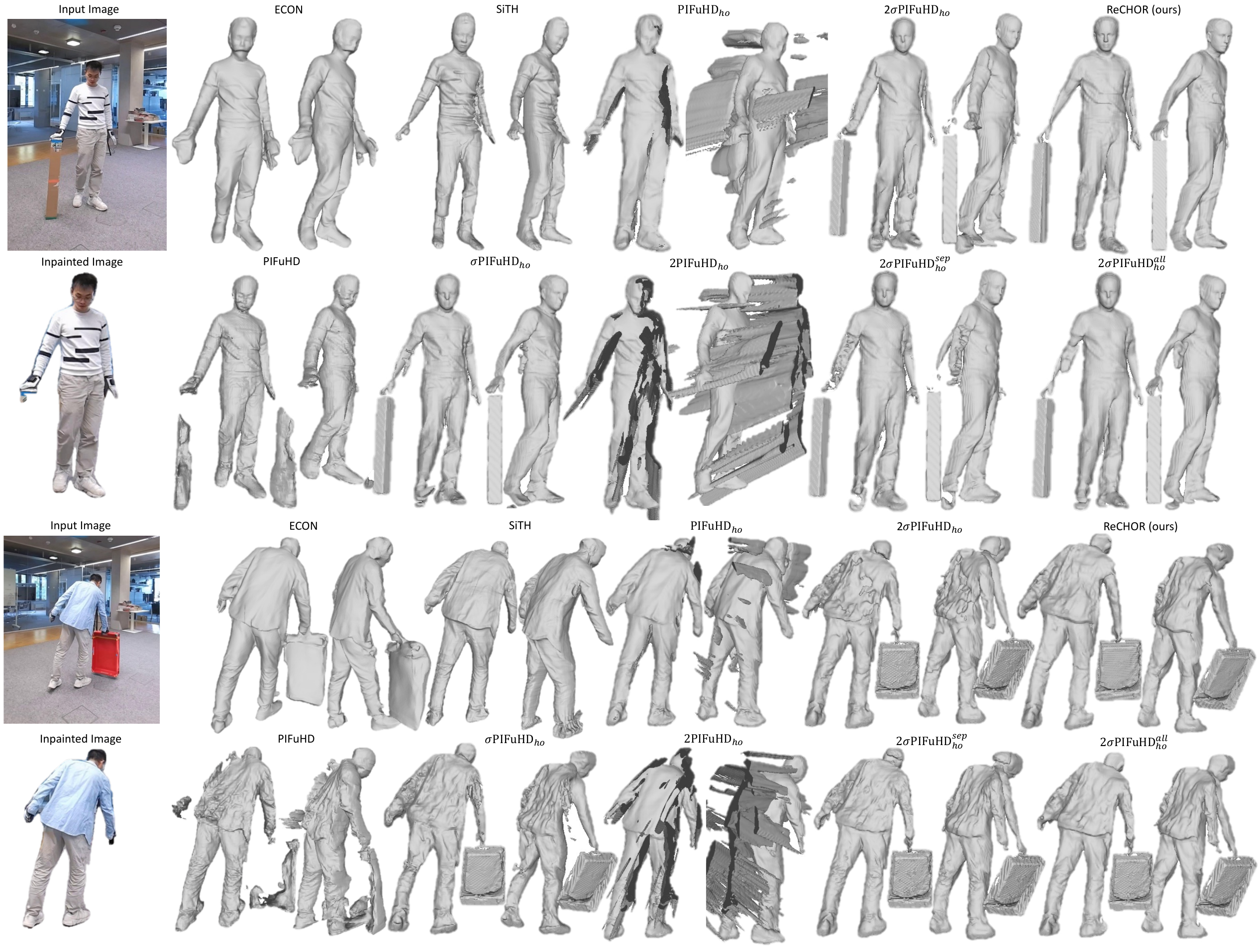}
\vspace{-6mm}
\caption{Additional visual results from BEHAVE~\cite{bhatnagar2022behave}. Examples obtained with methods that aim to reconstruct 3D humans as well as with the baselines considered in the main paper are shown. Front and side views are shown.}
\label{fig:behave_supp}
\end{figure*}

In this section, we present additional visual results of human-object  shapes reconstructed using \name, as well as comparisons with related methods that aim to reconstruct high-quality 3D humans (PIFuHD~\cite{pifuhd}, ECON~\cite{econ}, SiTH~\cite{ho2024sith}) and the baseline models introduced in~\cref{ssec:comp} of the main paper. 
Specifically,~\cref{fig:comp1_supp} presents results of the methods not included in~\cref{fig:comp1} of the main paper on \dataname dataset, while~\cref{fig:comp2_supp} illustrates reconstructed human-object shapes on the BEHAVE~\cite{bhatnagar2022behave} dataset that were omitted from~\cref{fig:comp2}.~\cref{fig:thuman_supp} and~\cref{fig:behave_supp} show new examples of reconstructed shapes using the considered methods on \dataname and BEHAVE dataset, respectively. 
Across all the presented figures, our approach reconstructs the most realistic human-object shapes with the fewest artefacts compared to related works. Human-focused reconstruction approaches cannot reconstruct objects, failing to achieve our goal of joint human-object reconstruction.
Even retraining PIFuHD on \dataname ($\mathrm{PIFuHD}_{\mathrm{ho}}$), incorporating  pose features ($\sigma\mathrm{PIFuHD}_{\mathrm{ho}}$,) or using two MLPs ($2\mathrm{PIFuHD}_{\mathrm{ho}}$), does not sufficiently improve results, with severe depth ambiguities in the reconstructed objects. 
The proposed baselines that leverage pose features along with two MLPs ($2\sigma\mathrm{PIFuHD}_{\mathrm{ho}}$, $2\sigma\mathrm{PIFuHD}^{all}_{\mathrm{ho}}$, $2\sigma\mathrm{PIFuHD}^{sep}_{\mathrm{ho}}$) can reconstruct objects but are significantly more prone to noise and artifacts compared to \name. The attention-based neural implicit model introduced by \name ensures the joint reconstruction of realistic clothed human and object shapes from single images, outperforming all related approaches.
\section{Discussion}
\label{sec:discussion}
\begin{figure*}[t]
  \centering
\includegraphics[width=\linewidth]{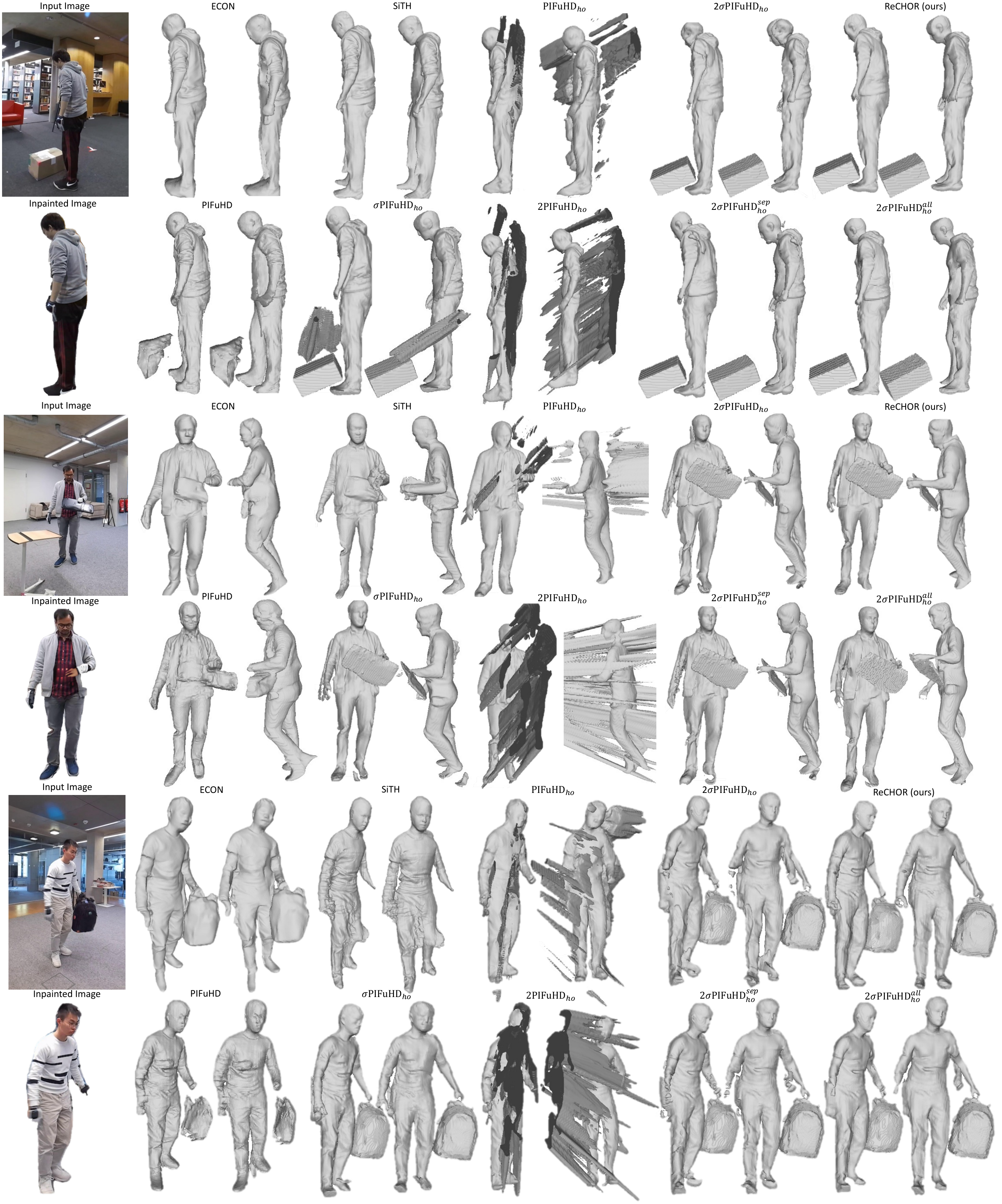}
\vspace{-6mm}
\caption{Visual comparisons from BEHAVE~\cite{bhatnagar2022behave} dataset with approaches that aim to reconstruct 3D humans as well as with baselines designed for fair comparisons. The same examples shown in~\cref{fig:comp2} are illustrated. Front and side views are shown.}
\label{fig:comp2_supp}
\end{figure*}

Existing works~\cite{econ, ho2024sith, pifuhd} for realistic human reconstruction are not designed to reconstruct objects. Consequently, when these methods are applied to real images containing both humans and objects, they behave differently in accordance with their network design, as illustrated in~\cref{fig:comp2_supp}. For instance, if these methods are conditioned on the SMPL model~\cite{ho2024sith,econ}, the object is not reconstructed unless it is very  close to the human. In such cases, these methods misinterpret the object as part of the clothing, merging it with the human mesh and resulting in a severe depth ambiguity. Similarly, methods that integrate normals~\cite{pifuhd} to reconstruct human shapes struggle to distinguish between clothing and objects, treating them as a single entity.
\\One potential solution is to crop the object out of the image and reconstruct only the human with the above methods. However, this requires reconstructing the object separately, without considering its spatial relationship with the human. In contrast, our approach is explicitly designed to jointly reason about humans and objects, distinguishing them as two distinct yet connected entities and reconstructing both in a shared 3D space.
\\We also highlight the novelty of our feature extraction and fusion strategy, where pixel-aligned features from each input are merged via transformer encoders. This design allows the model to jointly learn both global and local contextual information about the scene. Capturing global information improves the understanding of human-object spatial relationships, while local information  allows the reconstruction of realistic shapes.
\\Despite its strengths, our approach is prone to certain limitations.  First, the reconstruction quality of smaller body parts, such as fingers and hair, can be improved. This refinement will improve the modeling of human-object interactions and will be addressed in future work. In addition, increasing the quality of the ground-truth object shapes of our dataset will allow the network to learn finer details for the reconstructed object shapes, further increasing the realism of the reconstruction.
\\
Although \name can reconstruct arbitrary object shapes, it currently relies on object pose priors from a method~\cite{xie2022chore} that requires known object templates, limiting its generalization to objects without predefined templates. This limitation can be addressed in the future by integrating priors from recent template-free methods~\cite{xie2023template_free}. 
\\Our method relies on state-of-the-art works for human-object pose and normal map estimation. As a result, if these estimations are noisy, the noise propagates through the pipeline, causing artifacts in the reconstructed human shape.
Note that all the quantitative results on \dataname use ground-truth SMPL-H models and occlusion masks, while 2D normal maps are generated with pix2pixHD~\cite{wang2018high}  from the inpainted images.
\\Finally, a key direction for future work involves estimating the appearance of clothed human-object reconstructions to further enhance realism.

%% file: tables/notation.tex
\begin{tabular}{c|c}
\textbf{Symbol}                         & \textbf{Definition}                                                                                                                     \\ \hline
$I_f$                                   & Input RGB image                                                                                                                         \\ \hline
$I_o$                                   & Segmented object image                                                                                                                  \\ \hline
$I_p$                                   & Segmented partial human image                                                                                                           \\ \hline
$I_h$                                   & Generated full-body human image                                                                                                         \\ \hline
$S_N$                                   & Estimated 2D normal map                                                                                                                 \\ \hline
$\sigma_{\{h,o\}}$    & Human and object pose features                                                                                                          \\ \hline
$M_i$                                   & Mask of occluded human body regions                                                                                                     \\ \hline
$M_s$                                   & Mask of SMPL human model                                                                                                                \\ \hline
$M_p$                                   & Mask of partial human                                                                                                                   \\ \hline
$\varepsilon$                           & Variational Auto Encoder                                                                                                                \\ \hline
$\epsilon_{\theta}$                     & Diffusion U-Net                                                                                                                         \\ \hline
$\mathcal{D}$                           & Diffusion decoder                                                                                                                                 \\ \hline
$f_{\theta}$                            & Iterative denoising process of $\epsilon_{\theta}$                                                                                      \\ \hline
$\phi_{\{h,o,f\}}$                      & \begin{tabular}[c]{@{}c@{}}Feature extracted with stacked hourglass networks \\ from $I_h$, $I_o$ and $I_f$\end{tabular}                \\ \hline
$X_{\{h,o\}}$                           & \begin{tabular}[c]{@{}c@{}}3D points sampled in human \\ and object 3D ground-truth meshes   \end{tabular}                                                                         \\ \hline
$\Phi^{\{h,o\}}_{\{h,o,f\}}$           & \begin{tabular}[c]{@{}c@{}}Pixel-aligned human and object features \\ by projecting $X_h$ and $X_o$ in  $\phi_{\{h,o,f\}}$\end{tabular} \\ \hline
$A_{\{h,o\}}$                           & \begin{tabular}[c]{@{}c@{}}Transformer encoders to merge human \\ and object features with image features\end{tabular}                  \\ \hline
$\varphi_{\{h,o\}}$                     & Merged human and object features output of $A_{\{h,o\}}$                                                                                \\ \hline
$f_{\{h,o\}}$                           & Implicit functions (MLPs)                                                                                                               \\ \hline
$\hat{s}_{\{h,o\}}$ & Human and object implicit representations                                                                                              
\end{tabular}

%% file: tables/train_conf.tex
\begin{tabular}{c|ccccc}
                            & \multicolumn{5}{c}{\textbf{synHOR}}                                                          \\ \hline
Methods                     & P2S $\downarrow$ & CD $\downarrow$ & IoU $\uparrow$ & Normal $\uparrow$ & f-Score $\uparrow$ \\ \hline
$\mathrm{Sep}_{no}$                  & 23.36            & 27.22           & .8337          & .7523             & 34.38              \\
$\mathrm{Sep}$                       & 47.69            & 81.93           & .4564          & .5868             & 10.27              \\
\name (ours) & \textbf{14.23}   & \textbf{17.15}  & \textbf{.8902} & \textbf{.8241}    & \textbf{48.84}    
\end{tabular}

%% file: tables/inp.tex
\begin{tabular}{c|ccccc}
                            & \multicolumn{5}{c}{\textbf{synHOR}}                                                          \\ \hline
Methods                     & P2S $\downarrow$ & CD $\downarrow$ & IoU $\uparrow$ & Normal $\uparrow$ & f-Score $\uparrow$ \\ \hline
No inpainting               & 19.19            & 22.24           & .8679          & .8080             & 40.77              \\
MAT                         & 15.39            & 18.94           & .8841          & .8132             & 46.44              \\
Normal                      & 15.18            & 17.70           & .8860          & .8163             & 47.02              \\
Both                        & 14.91            & 18.49           & .8859          & .8160             & 48.26              \\
\name (ours) & \textbf{14.23}   & \textbf{17.15}  & \textbf{.8902} & \textbf{.8241}    & \textbf{48.84}    
\end{tabular}